%% file: main.tex
\documentclass[11pt]{article}

\usepackage[final]{acl}

\usepackage{times}
\usepackage{latexsym}

\usepackage[T1]{fontenc}

\usepackage[utf8]{inputenc}

\usepackage{microtype}

\usepackage{inconsolata}

\usepackage{graphicx}

\usepackage{booktabs}       
\usepackage{amsfonts}       
\usepackage{nicefrac}       
\usepackage{microtype}      
\usepackage{xcolor}         
\usepackage{amsmath} 
\usepackage{tikz}
\usetikzlibrary{arrows.meta}
\usepackage{multirow}
\usepackage{makecell}
\usepackage{algorithm}
\usepackage{algpseudocode}
\usepackage{makecell}

\usepackage{amsmath}
\usepackage{amssymb}
\usepackage{mathtools}
\usepackage{amsthm}
\usepackage{mathrsfs}
\usepackage{array}

\newcommand{\ModelName}{{DreamCUB}}
\newcommand{\E}[3][]{\mathbb{E}_{#2}#1[#3#1]}
\input{math_commands.tex}

\title{Dream to Chat: Model-based Reinforcement Learning on Dialogues with User Belief Modeling}

\author{
 \textbf{Yue Zhao\textsuperscript{1}},
 \textbf{Xiaoyu Wang\textsuperscript{1,2,\thanks{Internship at Geely}}},
 \textbf{Dan Wang\textsuperscript{1}},
 \textbf{Zhonglin Jiang\textsuperscript{1}},
 \textbf{Qingqing Gu\textsuperscript{1}},
 \\
 \textbf{Teng Chen\textsuperscript{1}},
 \textbf{Ningyuan Xi\textsuperscript{1,3,\footnotemark[1]}},
 \textbf{Jinxian Qu\textsuperscript{1}},
 \textbf{Yong Chen\textsuperscript{1}},
 \textbf{Luo Ji\textsuperscript{1}},
\\
\\
 \textsuperscript{1}Geely AI Lab,
 \textsuperscript{2}Beijing Institute of Technology,
 \textsuperscript{3} Beihang University,
\\
 \small{
   \textbf{Correspondence:} \href{Luo.Ji1@geely.com}{ Luo.Ji1@geely.com}
 }
}

\begin{document}
\maketitle
\begin{abstract}
World models have been widely utilized in robotics, gaming, and autonomous driving. However, their applications to natural language tasks are relatively limited. In this paper, we construct the dialogue world model, which could predict future utterances and user beliefs, including emotion, sentiment, and intention. In this paper, we propose a framework called DreamCUB, which shows that this user belief modeling and the entire dialogue world model can be established by LLM post-training. By defining a POMDP, we apply model-based reinforcement learning to the dialogue system and solve it by maximizing the information bottleneck. Experiments show that the pretrained dialogue world model can achieve state-of-the-art performances on emotion classification and sentiment identification, while dialogue quality is also enhanced by joint training of policy, critic and dialogue world model. Further analysis reveals that DreamCUB holds a reasonable exploration-exploitation balance and also transfers well to out-of-domain scenarios such as empathetic dialogues.
\end{abstract}

\section{Introduction}







%




Due to strong capabilities, modern Large Language models (LLM) have obtained remarkable progress on dialogue systems \cite{kang-etal-2024-large,zhou-etal-2024-think}. Among the training pipeline of conversational LLM, reinforcement learning from human feedback (RLHF) \cite{ouyangTrainingLanguageModels2022a} is an important post-training stage that bootstraps the human preference and achieves a deeper alignment by interactive sampling. Although PPO \cite{schulman2017proximal} is employed as the usual approach, its variants, such as DPO and GRPO, are also proposed to improve the dialogue policy. However, reinforcement learning (RL) is often subject to low sampling efficiency, high performance variance, and high computational overhead. When applied to dialogue systems, these issues become more challenging when the model size is large and the annotation is consuming.

\begin{figure}[t]
    \centering
    \includegraphics[width=\columnwidth]{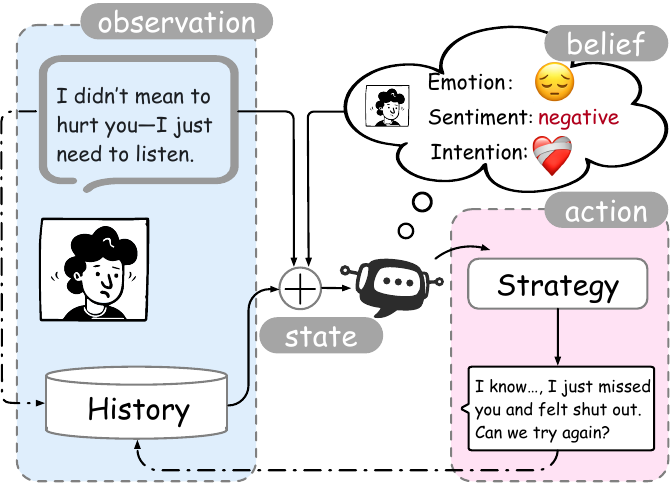}
    \caption{Paradigm of {\ModelName}, where we introduce \textbf{user belief modeling}, to speculate the unobservable state in dialogue. State becomes the union of observation and belief, which further enhances the policy.}
    \label{fig:paradigm}
\end{figure}

To alleviate these issues, Model-Based Reinforcement Learning (MBRL) \cite{suttonDynaIntegratedArchitecture1991,deisenrothPILCOModelbasedDataefficient2011} is proposed, which enables the agent to learn the environment model and use it to simulate, plan, and act. Combining with recent progress on World Models (WM) \cite{haWorldModels2018}, MBRL has been a powerful solution for visual control \cite{Hafner2020Dream}, game \cite{pmlr-v97-hafner19a}, auto-driving \cite{10547289} and also dialogue system \cite{peng-etal-2018-deep,xu-etal-2025-efficient}. For example, DDQ \cite{peng-etal-2018-deep} proposes the world model of dialogue which can predict the dialogue contents. Nevertheless, dialogues are highly sensitive to human psychological states, such as emotion and sentiment \cite{10.1145/3583780.3615265,10.5555/3545946.3598705}. People's reasoning, expression and intention can be affected and influenced by these inner states. However, such states are unobservable, while current MBRL studies on dialogues are based on observable states only, \textit{i.e.}, utterances. On the other hand, previous research on empathetic dialogue systems has mostly focused on generating responses given certain emotions. However, being empathetic not only requires responding based on self-emotions, but more importantly, calls for the understanding of user emotions and intentions, to respond appropriately \cite{lin-etal-2019-moel}.

To bridge these gaps, in this paper, we introduce the user belief modeling into the MBRL framework, to provide a more thorough understanding of the dialogue policy. Such user beliefs may include emotion, sentiment and intention, which are unobservable states for the agent, forming a Partially Observable Markov Process (POMDP). Correspondingly, our Dialogue World Model (DWM) can not only generate future dialogue utterances, but also recognize user beliefs and behave as the reward model. To solve this problem, we refer to the theoretical derivations of POMDP-based MBRL studies \cite{pmlr-v162-chen22q}, and deduce the DWM-RL algorithm based on the information bottleneck. Combining user belief modeling, DWM and MBRL, we propose the framework called \textbf{Dream} to \textbf{C}hat with \textbf{U}ser \textbf{B}elief (\textbf{\ModelName}). {\ModelName} simulates user belief and emotional dynamics over the course of interaction. Rather than relying on static emotion classification or purely supervised generation, \ModelName\ enables an agent to imagine possible future dialogue trajectories, reason about long-term emotional impact, and plan supportive responses accordingly. Figure \ref{fig:paradigm} illustrates the paradigm of {\ModelName}. We summarize our contributions as follows:

\begin{itemize}
    \item We redefine the Dialogue World Model which models user beliefs, to capture the sentimental and emotional dynamics.
    \item We introduce \textbf{\ModelName}, a model-based reinforcement learning framework to apply the knowledge of Dialogue World Model on dialogue systems.
    \item We empirically validate our approach on daily and empathetic dialogue datasets, showing accurate emotional predictions, high response quality and strong generalizations.
\end{itemize}

\section{Preliminaries}

\paragraph{POMDP.} A Partially Observable Markov Decision Process (POMDP) models the decision-making process under uncertainty when the system state is not fully observable. It is defined as 5-tuple:
\[
\mathcal{P} = (\mathcal{S}, \mathcal{A}, \mathcal{O}, \mathcal{T}, \mathcal{R})
\] 
where $\mathcal{S}$ is the state space, $\mathcal{A}$ is the action space, $\mathcal{O}$ is the observation space, $\mathcal T(s'|s, a)$ is the transition model, and $\mathcal R(s)$ is the reward function. 



\begin{figure}[t]
    \centering
    \includegraphics[width=\columnwidth]{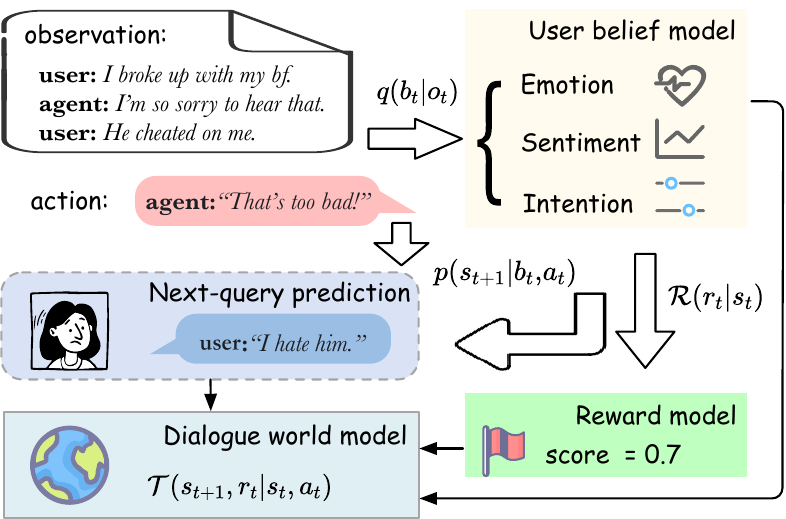}
    \caption{The dialogue world model (DWM) $\mathcal{T}(s_{t+1}, r_t | s_t, a_t)$ consists of three parts, the user belief model $q(b_t | o_t)$, the next-query prediction model $p(s_{t+1} | b_{t}, a_{t})$ and the reward model $\mathcal{R}(r_t|s_t)$.}
    \label{fig:dwm}
\end{figure}

\begin{figure*}[t]
    \centering
    \includegraphics[width=2.0\columnwidth]{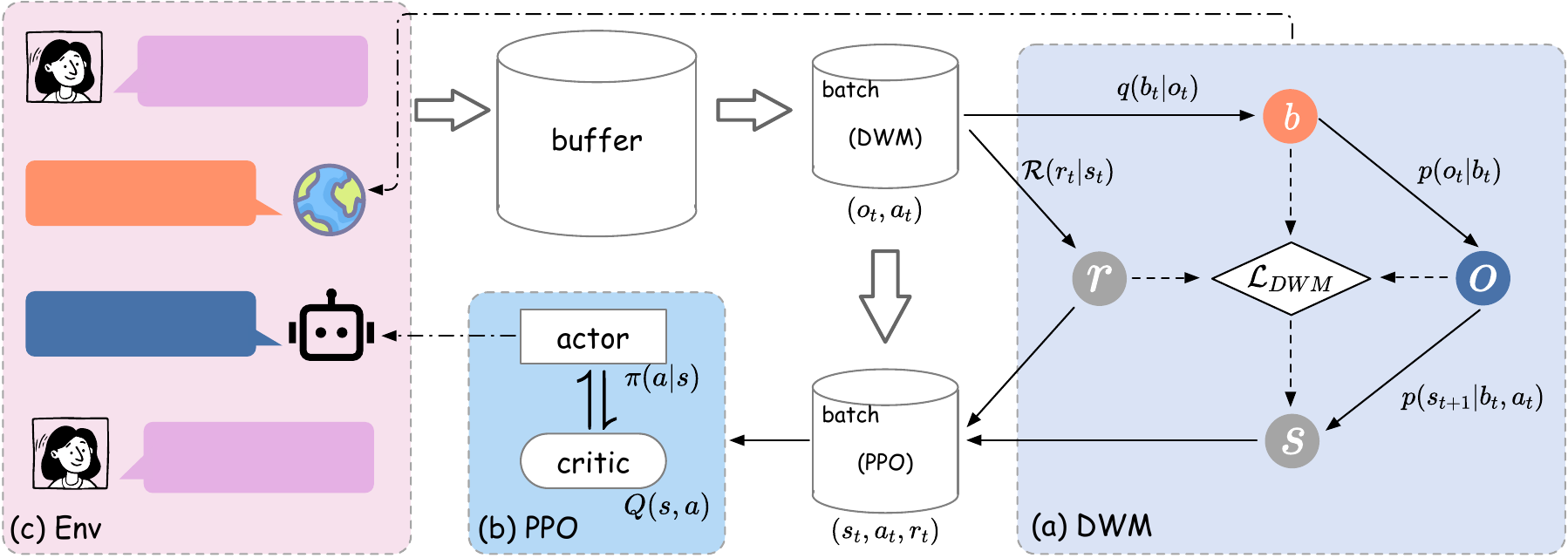}
    \caption{Training framework of {\ModelName}. (a) Dynamics learning of DWM. (b) Behavior Learning of dialogue policy. (c) Interaction with the environment.}
    \label{fig:framework}
\end{figure*}

\paragraph{Reward modeling.} Application of RL on textual environments requires Reward Model (RM) \citep{ouyangTrainingLanguageModels2022a}, which is trained from pairwise preference data $(x, y_{+}, y_{-})$ with $x$ as the input, $y_{+}$ and $y_{-}$ are positive and negative responses. RM is usually implemented by an LLM with the classification head added, which produces a 0-1 score. Its loss can be derived from human preference distribution by the Bradley-Terry \citep{bradley1952btmodel} model
\begin{equation}
\mathcal{L}_{\mathcal{R}} = \frac{1}{N} \sum_{i=1}^N \log \sigma ( \mathcal{R}(y_{+}^i|x^i)-\mathcal{R}(y_{-}^i|x^i)) \label{equation:rm}
\end{equation}
where $\mathcal{R}$ denotes RM, $\mathcal{L}$ is the loss, and $\sigma$ is the sigmoid function.

\paragraph{RLHF.} The generative policy on language tasks solves the following problem:
    \begin{equation}\max_{\pi_\theta}\mathbb{E}_{\vx\sim \mathcal{D}, \vy\sim \pi_\theta(\cdot|\vx)}\left[r_\phi(\vy|\vx)-\mathcal{L}_{KL}\right]\label{equation:rl_reward}\end{equation}
    where $\mathcal{L}_{KL}=\beta\KL(\pi_\theta(\cdot|\vx)\Vert \pi^{\text{SFT}}(\cdot|\vx))$ is the regularization term which prevents the RL policy from deviating from SFT too much. One usual solution is to employ PPO \citep{schulman2017proximal} to optimize the modified reward $r_\phi(\vy|\vx)-\beta\left(\log\pi_\theta(\vy|\vx)-\log \pi^{\text{SFT}}(\vy|\vx)\right)$.

\section{Method}

\paragraph{Tasks formulation.} Dialogue can be characterized by an interleaved sequence of user's $query$ and agent's $response$. At the $T$-th turn, we denote the dialogue history as
\begin{align}
    hist(T) := \{query(t), resp(t) \}_{0:T-1} 
\end{align}
where $hist$ and $resp$ abbreviate the history and response, respectively.

Recent studies usually bootstrap and annotate the agent's reply $strategy$, to have enhanced $response$ grounded by $strategy$. In this work, we further argue that the user's state, called $belief$, can also be modeled and behaves as the contextual information of subsequent $strategy$ and $response$. Such $belief$ may include the user's $emotion$, $sentiment$, and $intention$. In this formulation, the determination pipeline becomes
\begin{align}
    hist \oplus query \rightarrow belief \rightarrow strategy \rightarrow resp \notag
\end{align} 

\paragraph{System definition.} The above formulation suggests $query$, $resp$, $hist$ and $strategy$ are observable to the agent while the user's $emotion$, $sentiment$ and $intention$ are unobservable. The system can then be described as a 5-tuple POMDP $(\mathcal{O}, \mathcal{S}, \mathcal{A}, \mathcal{R}, \mathcal{T})$:


\noindent $\bullet$ Observation\: $o = (hist,query) \in \mathcal{O}$ \\
\noindent $\bullet$ Belief: $b = (emotion,sentiment,intention)$ \\
\noindent $\bullet$ State: $s = (o,b) \in \mathcal{S}$ \\
\noindent $\bullet$ Action: $a = (strategy,resp) \in \mathcal{A}$ \\
\noindent $\bullet$ Reward $r = \mathcal{R}(s)$ with $s$ as input instead of $o$\\
\noindent $\bullet$ Transition Function: $\mathcal{T}:= \mathcal{S} \times \mathcal{A} \rightarrow \mathcal{S}$. 


\paragraph{Model implementation.} To interpret this POMDP, we employ the model-based RL framework consisting of the following models:

\noindent $\bullet$ Belief inference model: $q(b_t | o_t)$\\ 
\noindent $\bullet$ Observation model: $p(o_{t} | b_t)$\\ 
\noindent $\bullet$ Belief Transition model: $p(b_{t+1} | b_{t}, a_{t})$\\ 
\noindent $\bullet$ Reward model: $\mathcal{R}(r_t | s_t)$\\
\noindent $\bullet$ Actor net: $\pi(a|s)$\\
\noindent $\bullet$ Critic net: $Q(s,a)$

Taking advantage of the strong linguistic capability of LLMs, we implement all the above models based on the foundation LLM, with the prompts in three categories:
\begin{enumerate}
    \item $q \leftarrow \text{LLM}(prompt_{cognitive})$: we implement the cognitive prompt \cite{wang-zhao-2024-metacognitive} for model $q$ which allows the identification of $emotion$, $sentiment$ and $intention$. 
    \item $p, \pi \leftarrow \text{LLM}(prompt_{generative})$: use generative prompts for $p(o_{t} | b_t)$, $p(b_{t+1} | b_{t}, a_{t})$ and the actor $\pi(a|s)$. 
    \item $\mathcal{R}, Q\leftarrow \text{LLM}(prompt_{classify}) \oplus \text{head}$: add the classification head on the last layer, which yields a 0-1 score \cite{ouyangTrainingLanguageModels2022a}. 
\end{enumerate}
with detailed prompt provided in Appendix \ref{sec:prompts}.

Specifically, we propose the term Dialogue World Model (\textbf{DWM}) $\mathcal{T}(s_{t+1}, r_t | s_{t}, a_{t})$ which contains three parts: the belief inference model $q(b_t | o_t)$ which is a cognitive model to identify the user belief; the belief transition model $p(s_{t+1} | b_{t}, a_{t}) = p(b_{t+1} | b_{t}, a_{t}) p(o_{t} | b_t)$ which conducts the next-query generation\footnote{In contrast, the dialogue policy $\pi(a|s)$ produces the next-response generation.}, and RM $\mathcal{R}(r_t|s_t)$ which produces the reward score. These three combined together, form the entire DWM. Figure \ref{fig:dwm} visualizes our DWM with more details.

\begin{algorithm*}
\caption{DWM-RL}  
\label{DWM-RL}
\begin{algorithmic}[1]
    \State Initialize the batch sizes $B_{DWM}$ and $B_{PPO}$, the window length $L$ and imagination horizon $H$
    \State Load pretrained cognitive model $q_{\xi}$, generative model $p_{\theta}$ and reward model $p_{\eta}(r_{\tau} | s_{\tau})$
    \State Initialize policy $\pi_{\phi}(a|s)$, critic $Q_{\psi}(s, a)$ and the buffer $\mathcal{B} = \{ \}$

    \State \textbf{while} not converged \textbf{do}:

            \State \Comment{\textit{Dynamic learning}}
            \State Draw $B_{DWM}$ data sequences $\{(o_t,a_t,r_t)\}_{t=k}^{k+L}$ from $\mathcal{B}$
            \State Inference belief state $q_{\xi}(b_t|o_t)$, rollout imaginary trajectories $\{(s_{\tau},a_{\tau})\}_{\tau=t}^{t+H}$ with  $p_{\theta}(s_{t+1} | b_t, a_t)$
            \State Update $\xi$, $\theta$ and $\eta$ by ELBO (Equation \ref{Eq:Jmodel}) 
            \State \Comment{\textit{Behavior learning}}
            \State Predict rewards $p_{\eta}(r_{\tau} | s_{\tau})$ for each $s_{\tau}$
                \State Draw $B_{RL}$ data sequences $\{(s_t,a_t,r_t)\}$ from $\{(s_{\tau},a_{\tau},r_{\tau})\}_{\tau=t}^{t+H}$
                \State Update $\phi$ and $\psi$ jointly by PPO (Equation \ref{equation:rl_reward})

        \State \Comment{\textit{Interact with the environment}}
        \State Get original query $o_1$ from dataset.
        \For{t = $1, \dots, T$}
        \State Inference the belief $b_t \sim q_{\xi}(b_t | o_t)$, forming the state $s_t = (o_t, b_t)$
        \State Determine the action $a_t \sim \pi_{\phi}(a_t | s_t)$
        \State Execute $a_t$ and get $o_{t+1}$, $r_t$
        \EndFor

    \State Add experience to buffer $\mathcal{B} = \mathcal{B} \cup \{(s_t,a_t,r_t)\}_{t=1}^{T}$ 
    \State \textbf {end while}

\end{algorithmic}
\end{algorithm*}

\paragraph{Algorithm.} Posterior of beliefs and rewards, given observations and actions, can be maximized jointly by the variational information bottleneck \citep{tishby2000ib}, or the Evidence Lower Bound (ELBO) \citep{jordan1999viintro}:

\newcommand{\eqbr}{\,\hookleftarrow\\[-1ex]}  
\newlength\widthE
\newcommand{\Ebelow}[3][]{\settowidth\widthE{$\operatorname{E}$} \mathop{\mathbb{E}}_{\vphantom{|^|}\mathmakebox[0.5\widthE][l]{#2}}#1[#3#1]}

{\small\begin{multline}
\log p(o_{1:T}, r_{1:T}|a_{1:T}) \\
\begin{aligned}
&\geq\sum_{t=1}^T \Big(
  \E{q(b_t|o_{\leq t},a_{<t})}{\log p(o_t|b_t) + \log \mathcal{R}(r_t | b_t)} \quad \\
  &\quad - \Ebelow[\big]{q(b_{t-1}|o_{t-1})}{D_{\mathrm{KL}}(q(b_t|o_t) \| p(b_t|b_{t-1},a_{t-1})}) \Big) \doteq \mathcal{L}_{\mathrm{DWM}}
\end{aligned}
\label{Eq:Jmodel}
\end{multline}}
\noindent with precise derivation in Appendix \ref{sec:append-ELBO}. This lower bound was originally proved by \cite{pmlr-v162-chen22q} which derives the following theorems:
\begin{theorem}
\label{elbo-thm}
The approximation error of the log-likelihood when maximizing the  $\mathcal{L}_{\mathrm{DWM}}$ (the derived ELBO) defined in Equation \ref{Eq:Jmodel} is:


\begin{equation}
\begin{aligned}\label{eq:elbo_error}
    &\log p(o_{1:T}, r_{1:T}|a_{1:T}) - \mathcal{L}_{\mathrm{DWM}} \\
    &= \Ebelow[\big]{q(b_{1:T}|o_{1:T},a_{1:T-1})}{
    \Sigma_{t=1}^{T}
    D_{\mathrm{KL}}(q(b_t|o_t) \| \bar{p}(b_t|o_t))}
\end{aligned}
\end{equation}
where $\bar{p}(b_t|o_t)$ denotes the true states. 
\end{theorem}
Based on the aforementioned consideration, we propose Algorithm \ref{DWM-RL}, the Dialogue World Model-based Reinforcement Learning (DWM-RL), which contains three stages: (i) Dynamic learning, (ii) Behavior learning and (iii) Interact with the environment. Figure \ref{fig:framework} shows the entire framework.

\begin{table*}[htbp!]
  \centering
  \resizebox{0.99\textwidth}{!}{
    \begin{tabular}{l|cc|cc|cc|c|c|cc|cc}
    \toprule
    \multicolumn{1}{c|}{\textbf{task $\rightarrow$}} & \multicolumn{6}{c|}{\textbf{sentiment classification }} & \multicolumn{2}{c|}{\textbf{intensity regression}} & \multicolumn{4}{c}{\textbf{emotion classification}} \\
    \toprule
    \multicolumn{1}{c|}{\multirow{2}[4]{*}{\textbf{model} $\downarrow$}} & \multicolumn{2}{c|}{\textbf{Amazon}} & \multicolumn{2}{c|}{\textbf{IMDb}} & \multicolumn{2}{c|}{\textbf{Yelp}} & \multicolumn{1}{c|}{\textbf{V-reg}} & \multicolumn{1}{c|}{\textbf{SST}} & \multicolumn{2}{c|}{\textbf{GoEmotion}} & \multicolumn{2}{c}{\textbf{E-c}} \\
\cmidrule{2-7}    \cmidrule{8-9}   \cmidrule{10-13}        
& \multicolumn{1}{c}{\textbf{ACC}} & \multicolumn{1}{c|}{\textbf{MaF1}} & \multicolumn{1}{c}{\textbf{ACC}} & \multicolumn{1}{c|}{\textbf{MaF1}} & \multicolumn{1}{c}{\textbf{ACC}} & \multicolumn{1}{c|}{\textbf{MaF1}} & \multicolumn{1}{c|}{\textbf{pcc}} & \multicolumn{1}{c|}{\textbf{pcc}} & \multicolumn{1}{c}{\textbf{ACC}} & \multicolumn{1}{c|}{\textbf{MaF1}} & \multicolumn{1}{c}{\textbf{MiF1}} & \multicolumn{1}{c}{\textbf{MaF1}} \\
    \midrule
    \textit{llama2-7b-chat} & 64.19  & 69.17  & 83.23  & 86.36  & 87.69  & 89.48 & 9.12  & 72.83  & 35.71  & 27.15   & 41.40  & 28.60 \\
    \multicolumn{1}{p{10em}|}{Emollama-chat-7b} & 56.95  & 63.43  & 73.52  & 82.90  & 74.46  & 81.01  & \bf 88.00  & 82.00  & 37.00  & \textbf{39.00 }  & \bf 69.30  & \bf 54.00 \\
    \bf DWM & \textbf{74.13 } & \textbf{73.89 } & \textbf{96.38 } & \textbf{96.38 } & \textbf{97.42 } & \textbf{97.31 }  & 86.38  & \textbf{90.28 } & \textbf{39.44 } & 30.41     & 51.32 & 48.67 \\
    \midrule
    \textit{llama2-13b-chat} & 69.54  & 71.93  & 90.66  & 91.51  & 90.07  & 91.06  & 24.06  & 81.10  & 27.80  & 33.70   & 42.40 & 30.20  \\
    \multicolumn{1}{l|}{Emollama-chat-13b} & 65.01  & 69.61  & 55.70  & 69.51  & 51.28  & 59.86  & \bf 88.40  & 81.60  & 35.00  & \bf 37.00 & \bf 69.60 & 54.50 \\
    \bf DWM & \bf 73.84 & \bf 73.68 & \bf 96.69 & \bf 96.69 & \bf 97.53 & \bf 97.41 & 88.36 & \bf 90.66 & \bf 37.21 & 33.81 & 69.41 & \bf 57.73 \\
    \midrule
    \multicolumn{1}{l|}{\textit{llama3-8b-instruct}} & 72.38  & 73.92  & 92.63 & 92.66 & 93.21  & 92.94  & 57.04  & 82.17  & 32.83  & \bf 34.43    & 43.95 & 41.38 \\
    \bf DWM ($q(b|o)$) & \textbf{87.87 } & \textbf{87.87 } & \textbf{96.99 } & \textbf{96.99 } & \bf 96.34 & \bf 96.17  & \bf 86.50  & \bf 90.19  & \bf 33.60  & 32.52   & \bf 58.39 & \bf 59.42  \\
    \bottomrule
    \end{tabular}%
    }
  \caption{Performance of dialogue world model compared with state-of-the-art emotional cognition models. V-reg and E-c are two subtasks of SemEval 2018 Task1. pcc denotes the Pearson correlation coefficient.}
  \label{tab:DWB_acc}%
\end{table*}%

\begin{table}[h] 
    \centering
    \renewcommand{\arraystretch}{1.2} 
    \resizebox{\columnwidth}{!}{ 
    \begin{tabular}{>{\centering\arraybackslash}m{1.5em}|c|c}
        \Xhline{2\arrayrulewidth}
        \multirow{3}{*}{\rotatebox[origin=c]{90}{\textbf{{history}}}} 
        & \textit{user:} & \textit{Did you hear about the robbery?}  \\
        \cline{2-3}
        & \textit{agent:} & \textit{Did I hear about it? I saw it happen.}   \\
        \cline{2-3}
        & \textit{user:} & \textit{Are you serious?}   \\ 
        \Xhline{1\arrayrulewidth}
        \multirow{2}{*}{\rotatebox[origin=c]{90}{\textbf{{belief}}}} 
        & \multicolumn{2}{l}{\textbf{Emotion: \textit{"surprise"}, Sentiment:\textit{"negative"}, \textit{"0.388"}}}  \\   
        \cline{2-3}
        & \textcolor{brown}{Ground Truth} & \textcolor{brown}{surprise, negative} \\
        \Xhline{2\arrayrulewidth}
        & \textit{agent:} & \textit{<inform> I was there.}   \\ 
        \midrule
        \multirow{2}{*}{\rotatebox[origin=c]{90}{\textbf{{query}}}} 
        & \multirow{2}{*}{\textit{user:}} & Predicted: \textbf{What went down?}  \\ 
        \cline{3-3}
        &   & Ground Truth: \textcolor{brown}{What happened ?} \\
        \Xhline{2\arrayrulewidth}
    \end{tabular}
    }
    \caption{Case of DWM on user belief cognition ($q(b_t|o_t)$) and next-query prediction ($p(o_{t} | b_t, o_{t-1})$). Contents from the original dataset are \textit{italic}, and results of DWM are \textbf{bolded}.}
    \label{tab:case_q}
\end{table}

\begin{table*}[htbp!]
\centering
\small
\begin{tabular}{l | ccc | ccc | ccc}
    \toprule
    \multicolumn{1}{c|}{\multirow{2}[2]{*}{Method}} &  \multicolumn{3}{c|}{Emotion} &  \multicolumn{3}{c|}{Strategy} &  \multicolumn{3}{c}{Response} \\ 
    \cmidrule{2-4} \cmidrule{5-7} \cmidrule{8-10}
    & ACC & MaF1 & $bias \downarrow$ & ACC & MaF1 & $bias \downarrow$ & B-2 & R-L & D-2 \\ 
    \toprule
    Direct & - & - & - & 52.60 & 18.03 & 1.66 & 3.35 & 10.33 & 44.74 \\
    \;+ Retrieve & - & - & - & 30.92  & 21.17  & 0.67  & 2.78 & 9.67 & 40.60 \\
    \;+ Refine & - & - & - & 48.27  & 28.28  & 0.70  & 2.56 & 8.70 & 43.67 \\
    \;+ Self-Refine & - & - & - & 49.76  & 22.15  & 1.18  & 2.40 & 7.75 & 34.01 \\
    \;+ CoT & - & - & - & 38.94  & 29.99  & \bf 0.27 & 1.78 & 6.00 & \bf 55.26 \\
    \;+ FSM & 73.01 & \underline{24.50} & \underline{1.63} & 46.86 & 21.22 & 1.30 & 2.70 & 9.44 & 38.75 \\
    \midrule
    \;+ SFT & 76.76 & 14.35 & 2.03 & 60.19 & 44.82 & 0.82 & \underline{6.81} & 18.52 & 43.36 \\
    \;+ CoT + SFT & \underline{83.48} & {15.60} & 1.98 & 60.11 & 44.90 & 0.66 & 6.61 & 18.07 & 42.87 \\
    \;+ FSM + SFT & 83.28 & 14.44 & 2.22 & \underline{64.05} & \underline{48.36} & 0.62 & 5.85 & \underline{21.77} & 47.43 \\
    \bf \;+ {\ModelName} (ours) & \bf 88.05 & \bf 50.88 & \bf 0.74 & \bf 67.80 & \bf 62.29 & \underline{0.33} & \bf 11.65 & \bf 29.09 & \underline{49.36} \\
    \bottomrule
\end{tabular}
\caption{ID results on automatic metrics on DailyDialog, including classification metrics such as Accuracy (ACC), Macro-F1 (MaF1) and $bias$, and generation metrics such as BLEU-2 (B-2), ROUGE-L (R-L) and Distinct-2 (D-2). The best results of each LLM are \textbf{bolded} and the second best are \underline{underlined}. 
}
\label{tab:ID_results}
\end{table*}

\section{Experiment}

\subsection{Settings}

\paragraph{Implementation.} Llama3.1-8B-Instruct \cite{llama3modelcard} is employed as the base model. Training is conducted on OpenRLHF \cite{hu2024openrlhf} with $L=1024$, $H=16$, $B_{DWM}=256$, $B_{PPO}=512$, $\gamma=0.9$, $\beta=0.01$. The learning rate is $5.0e-7$, training epoch is 1 and the replay buffer size is 24,000. RM is trained with positive responses from the original dataset and negative responses from dynamic sampling.



\paragraph{Datasets.} For DWM pretraining, we employ three types of tasks:
\begin{enumerate}
    \item Sentiment classification: classify either Positive or Negative from the user query. We use Amazon\footnote{\url{http://jmcauley.ucsd.edu/data/amazon/}}, Yelp\footnote{\url{https://www.yelp.com/dataset/download}}, and IMDB \cite{maas-EtAl:2011:ACL-HLT2011} as benchmarks.
    \item Sentiment intensity regression: predict a 0-1 score indicating the user's sentiment polarity\footnote{0 means fully negative and 1 means fully positive.}. We use Stanford Sentiment Treebank (SST) \cite{socher-etal-2013-recursive} and the corresponding subtask in SemEval-2018 Task1: Affect in Tweet \cite{mohammad-kiritchenko-2018-understanding}.
    \item Emotion classification: select the appropriate emotion from the candidates, such as joy, anger, sad, etc. We use GoEmotion \cite{demszky-etal-2020-goemotions} and again the corresponding subtask in SemEval-2018 \cite{mohammad-kiritchenko-2018-understanding}.
\end{enumerate}

For PPO training, we use DailyDialog \cite{li-etal-2017-dailydialog}, ESconv \cite{liu2021ESconv}, EmpatheticDialogues \cite{rashkin-etal-2019-towards}. The first two have annotations of emotion, strategy and response, while the last one only has annotations of emotion and response. To gain significant generalizability, we use DailyDialog \cite{li-etal-2017-dailydialog}, which is focused on daily topics, as both training and in-domain (ID) test sets. The other two, which are more focused on empathetic dialogue, are used for out-of-domain (OOD) evaluation purposes only.

\paragraph{Metrics.} For classification tasks, we employ the metrics of accuracy (ACC), Micro-F1 (MiF1) and Macro-F1 (MaF1). We also refer to the evaluation methods proposed by \citet{kang-etal-2024-large}, which propose the $bias$ based on the Bradley-Terry model~\citep{bradley1952btmodel}. A smaller $bias$ means less bias, therefore it is better. For regression tasks, we use the Pearson correlation coefficient (pcc). For the generation task, we utilize the BLEU-2 (B-2), Rouge-L (R-L) and Distinct-2 (D-2). The first two are similarity-based metrics, while the last one encourages response diversity. We also conduct human annotations to evaluate the responses. We leave the annotation principle and metric details in the Appendix.

\subsection{Training of {\ModelName}} 
\begin{figure}[htbp!]
    \centering
    \includegraphics[width=0.45\linewidth]{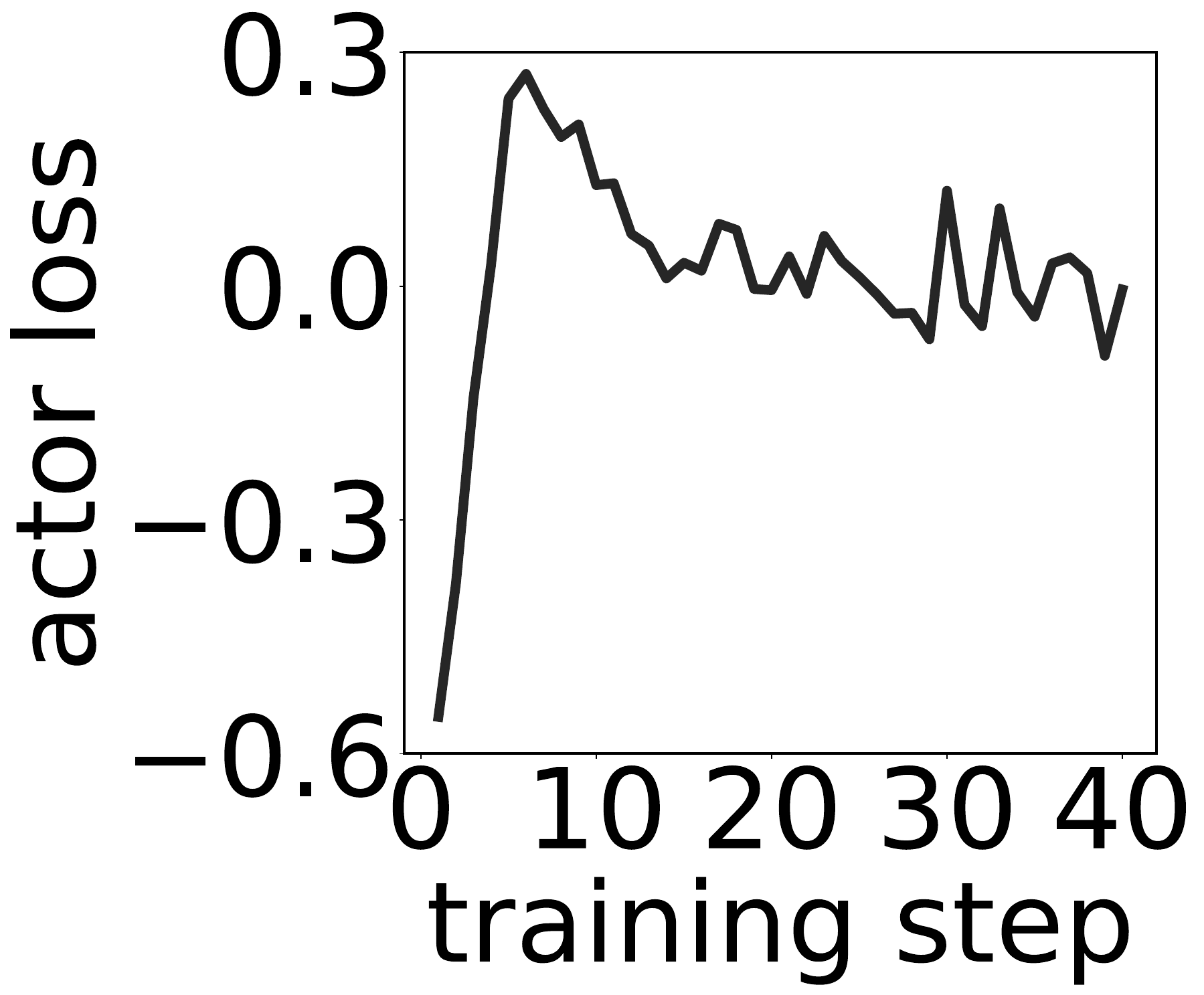}
    \hspace{0.1in}
    \includegraphics[width=0.45\linewidth]{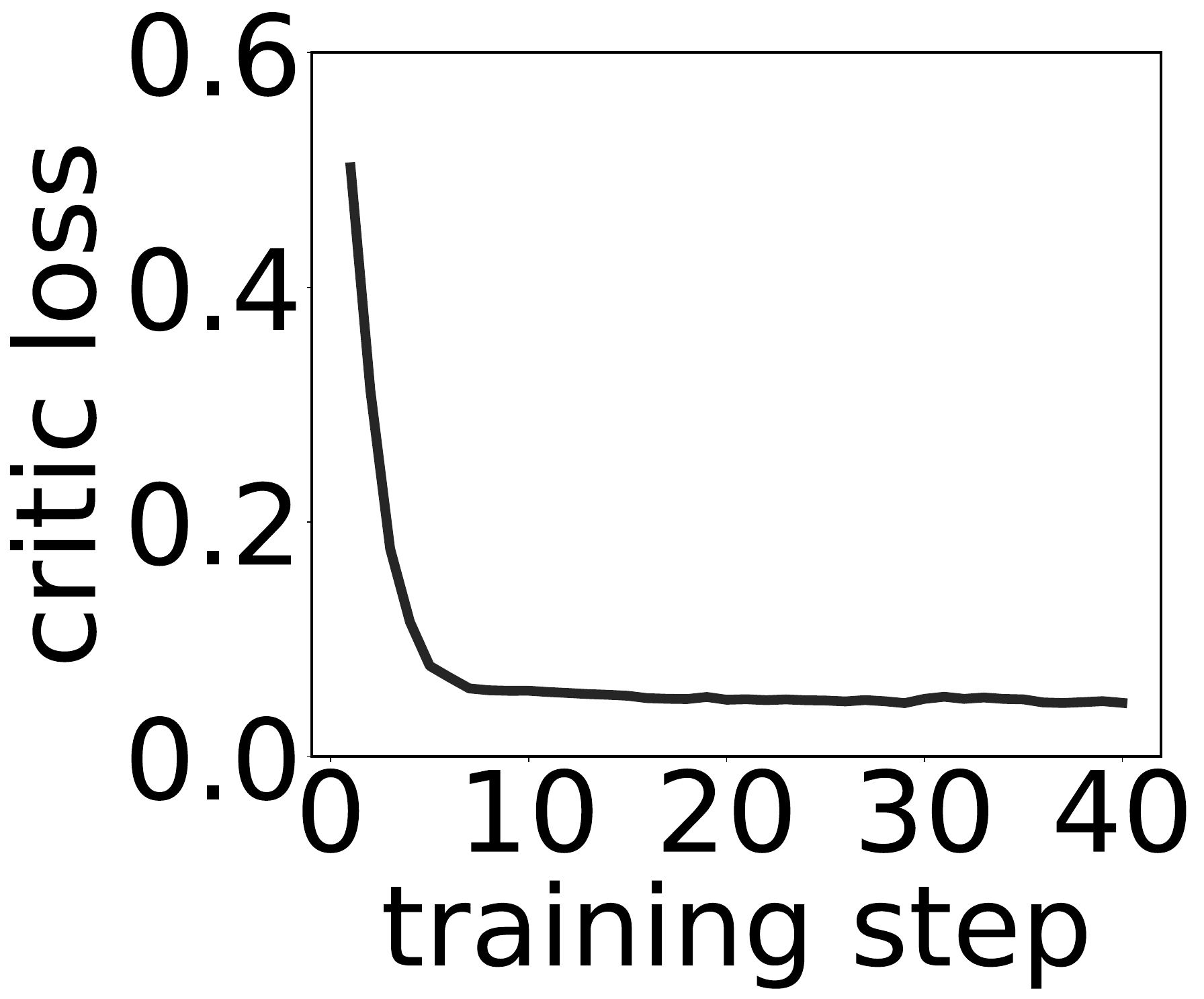}
    \includegraphics[width=0.45\linewidth]{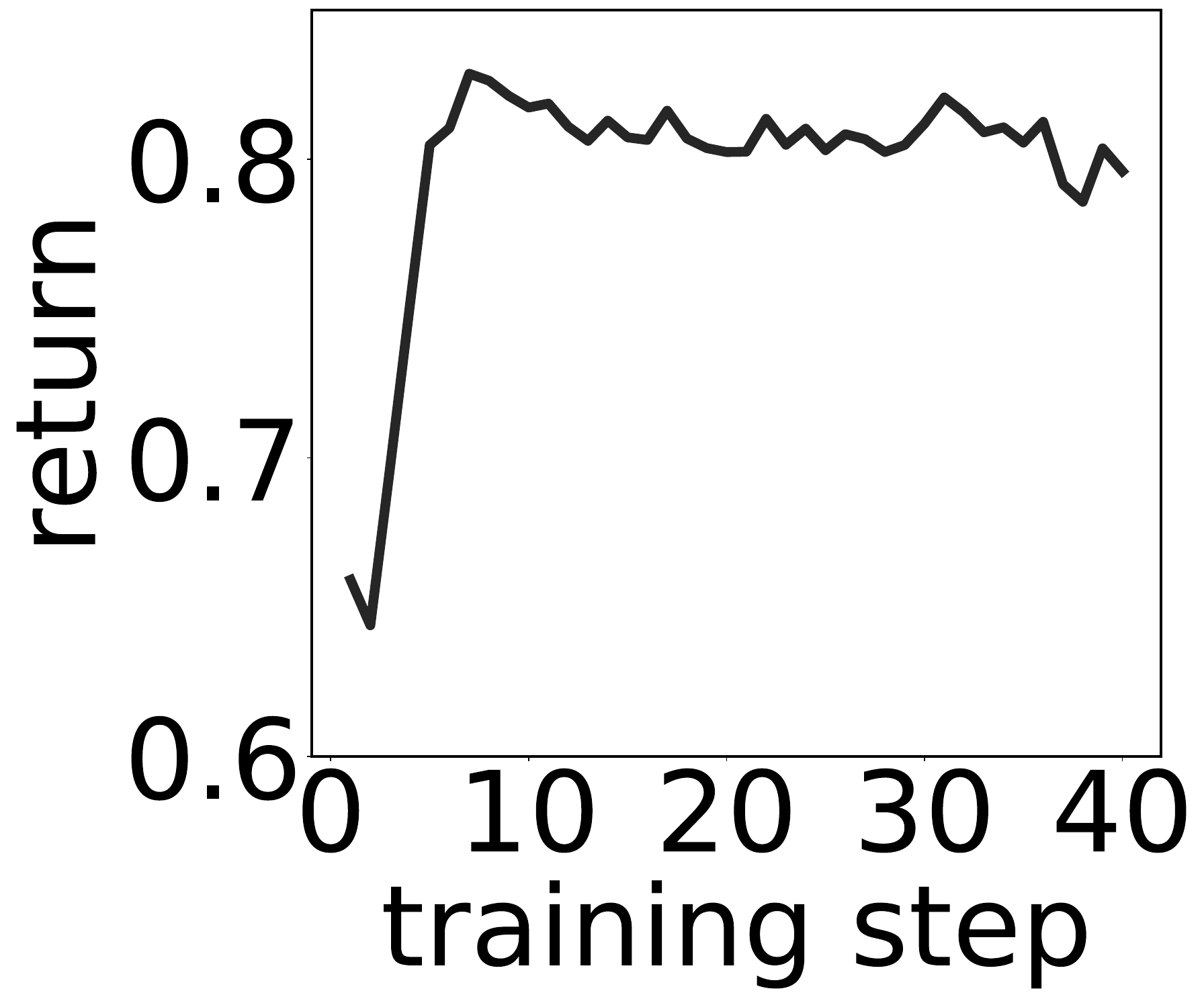}
    \hspace{0.1in}
    \includegraphics[width=0.45\linewidth]{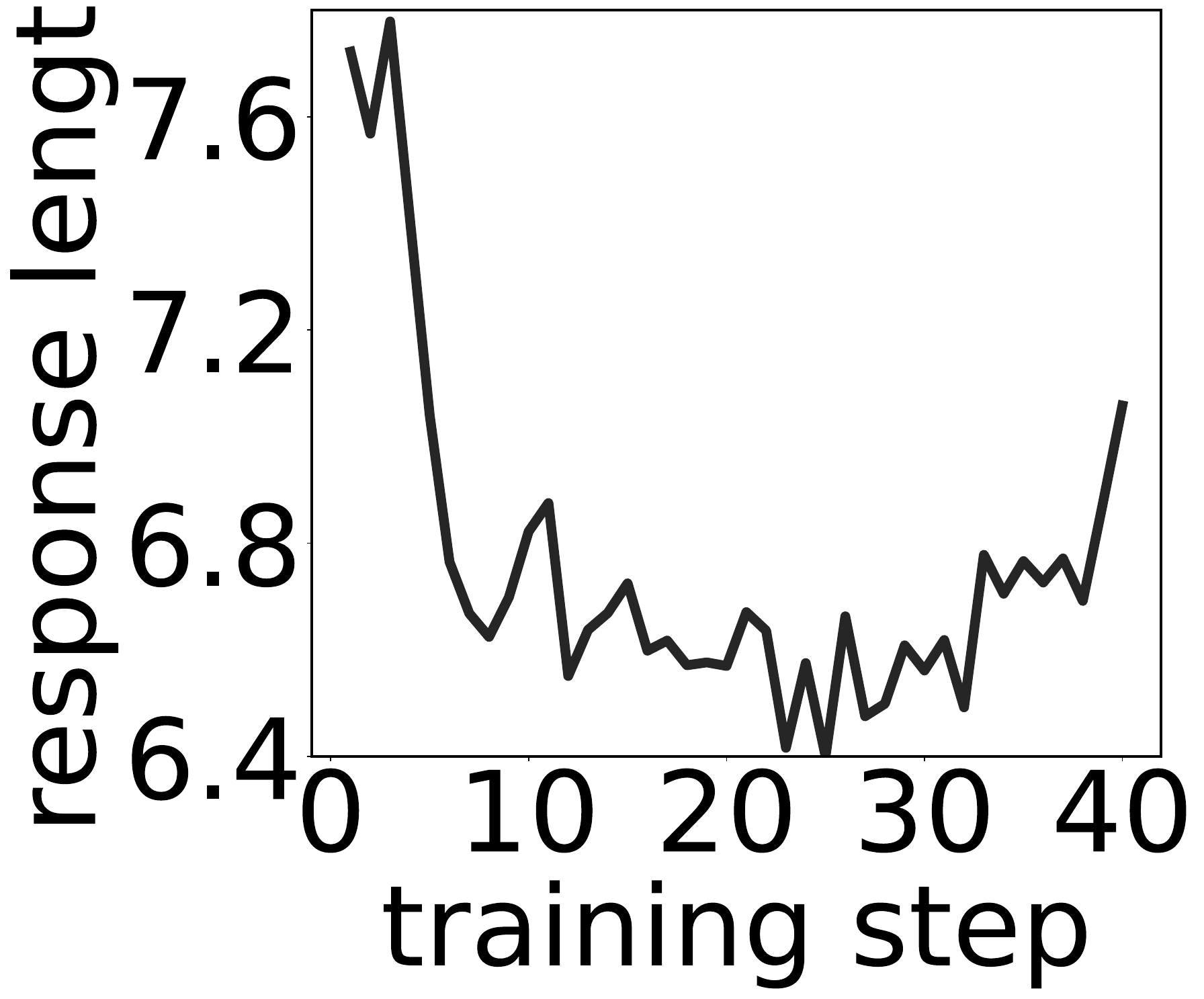}
    \caption{Training plots of {\ModelName}, including the actor loss (top-left), the critic loss (top-right), return (bottom-left) and reward (bottom-right).}
    \label{fig:training_curves}
\end{figure}

Figure \ref{fig:training_curves} visualizes the training curves, which shows that our Algorithm \ref{DWM-RL} converges and the return can be maximized. More specifically, Figure \ref{fig:training_curves} (bottom-right) highlights a preference evolution of the dialogue policy, the response length. At the beginning of training, the LLM tends to provide long responses, which are not natural enough considering the daily conversation situation. As joint training with DWM, the responses start to become shorter and finally reach a balance.

\subsection{Results of dialogue world model}

\paragraph{Emotion Cognition.} Table \ref{tab:DWB_acc} shows our DWM after the pretraining. We achieve state-of-the-art accuracy on all three types of emotional cognitive tasks, surpassing the base model and EmoLLama. To be consistent with our RL training, we use the Llama3-based version for the subsequent formal experiments.
Table \ref{tab:case_q} shows a good case of emotion cognition.

\paragraph{Dialogue Generation.} Our system transition model ($p$) of DWM needs to predict the user intention or query, based on the current conversation context. However, next-query prediction is difficult to have qualitative results, since user queries could be open-ended topics. Instead, Table \ref{tab:case_q} shows a typical case of $p$. One can observe that $p$ can understand contextual information and generate reasonable user queries that sometimes are similar to the ground truth.

\paragraph{Scalability.} Table \ref{tab:DWB_acc} also shows results of the 13B-based experiment, in which our DWM still performs better than the base model and EmoLlama on most of the metrics, suggesting our method is scalable to higher model and data sizes.



\begin{table*}[htbp!]
\centering
\small
\begin{tabular}{l l | ccc | ccc | ccc}
    \toprule
    & \multicolumn{1}{c|}{\multirow{2}[2]{*}{Method}} &  \multicolumn{3}{c|}{Emotion} &  \multicolumn{3}{c|}{Strategy} &  \multicolumn{3}{c}{Response} \\ 
    \cmidrule{3-5} \cmidrule{6-8} \cmidrule{9-11}
    & & ACC & MaF1 & $bias \downarrow$ & ACC & MaF1 & $bias \downarrow$ & B-2 & R-L & D-2 \\ 
    \toprule
    \parbox[t]{5mm}{\multirow{4}{*}{\rotatebox[origin=c]{90}{\parbox[c]{1.2cm}{\centering \scriptsize ESconv}}}}
    & \;SFT & 25.12 & 11.38 & 2.65 & 11.15 & 5.54 & 2.19 & \underline{3.30} & \underline{12.90} & 27.67 \\
    & \;CoT + SFT & \underline{32.90} & \bf 15.48 & \underline{2.21} & 15.28 & 8.09 & \bf 1.75 & 2.33 & 9.00 & \underline{31.13} \\
    & \;FSM + SFT & 30.23 & 6.84 & 2.62 & \underline{18.76} & \underline{8.12} & 1.88 & 2.70 & 10.46 & 28.10 \\
    & \bf \;{\ModelName} (ours) & \bf 34.26 & \underline{14.78} & \bf 1.94 & \bf 30.78 & \bf 10.90 & \underline{1.80} & \bf 3.68 & \bf 13.71 & \bf 33.23 \\
    \toprule
    \parbox[t]{5mm}{\multirow{4}{*}{\rotatebox[origin=c]{90}{\parbox[c]{1.2cm}{\centering \scriptsize Empathetic \\ -Dialogues}}}}
    & \;SFT & 4.03 & 1.44 & 5.44 & N/A & N/A & N/A & 2.56 & 7.68 & 34.83 \\
    & \;CoT + SFT & \underline{12.20} & \underline{7.77} & \bf 3.60 & N/A & N/A & N/A & 2.56 & 9.81 & \bf 39.39 \\
    & \;FSM + SFT & 4.59 & 2.20 & 5.57 & N/A & N/A & N/A & \underline{2.61} & \underline{9.87} & 30.52 \\
    & \bf \;{\ModelName} (ours) & \bf 16.49 & \bf 17.58 & \underline{5.15} & N/A & N/A & N/A & \bf 4.03 & \bf 13.15 & \underline{37.08} \\
    \bottomrule
\end{tabular}
\caption{OOD results on automatic metrics on ESconv and EmpatheticDialogues, including classification metrics such as Accuracy (ACC), Macro-F1 (MaF1) and $bias$, and generation metrics such as BLEU-2 (B-2), ROUGE-L (R-L) and Distinct-2 (D-2). The best results of each LLMs are \textbf{bolded} and the second best are \underline{underlined}. 
}
\label{tab:OOD_results}
\end{table*}

\begin{table*}[htbp!]
\centering
\small
\begin{tabular}{l|ccccccc}
    \toprule
 \multicolumn{1}{c|}{Method} & Fluency  & Emotion & Acceptance & Effectiveness & Sensitivity & Alignment & Satisfaction \\ 
    \toprule   
    Direct  & 2.95$\pm$1.41 & 3.00$\pm$1.34 & 2.60$\pm$1.15 & 2.40$\pm$0.92 & 2.70$\pm$1.08 & 2.70$\pm$1.08 & 2.60$\pm$1.41 \\
    \;+ Refine & 3.09$\pm$1.25 & 3.09$\pm$1.16 & 2.73$\pm$1.22 & 2.91$\pm$1.41 & 2.91$\pm$1.23 & 2.82$\pm$1.25 & 2.84$\pm$1.40 \\
    \;+ Self-Refine & 3.10$\pm$1.29 & 3.15$\pm$1.38 & 2.80$\pm$1.19 & 2.70$\pm$1.14 & 2.90$\pm$1.03 & 2.80$\pm$1.16 & 2.80$\pm$1.20 \\
    \;+ CoT  & 3.08$\pm$1.02 & 3.08$\pm$1.29 & 2.83$\pm$1.27 & 2.67$\pm$1.06 & 3.00$\pm$1.27 & 2.83$\pm$1.13 & 2.83$\pm$1.10 \\
    \;+ FSM  & 3.30$\pm$1.32 & 3.35$\pm$1.38 & 2.90$\pm$1.17 & 2.90$\pm$1.03 & 3.00$\pm$1.46 & 2.90$\pm$1.15 & 2.93$\pm$1.19 \\
    \midrule
    \;+ SFT & 3.15$\pm$1.44 & 3.40$\pm$1.30 & 2.70$\pm$1.19 & 2.70$\pm$1.20 & 2.90$\pm$1.24 & 3.30$\pm$1.32 & 2.90$\pm$1.32 \\
    \;+ CoT + SFT  & 3.67$\pm$1.21 & \textbf{3.61}$\pm$1.17 & 3.22$\pm$1.25 & 3.67$\pm$1.26 & 3.56$\pm$1.13 & 3.35$\pm$1.39 & 3.45$\pm$1.31 \\
    \;+ FSM + SFT  & 3.80$\pm$1.26 & 3.55$\pm$1.16 & 3.40$\pm$1.21 & 3.70$\pm$1.14 & 3.80$\pm$1.06 & 3.70$\pm$1.04 & 3.65$\pm$1.19 \\
    \bf \;+ {\ModelName} & \textbf{3.85}$\pm$1.10 & 3.52$\pm$1.47 & \textbf{4.09}$\pm$0.87 & \textbf{3.90}$\pm$0.99 & \textbf{3.86}$\pm$1.14 & \textbf{4.01}$\pm$1.09 & \textbf{3.98}$\pm$1.12 \\
    \bottomrule
    \end{tabular}%
\caption{Average human scores (with standard deviations) of response quality on ESconv and EmpatheticDialogues.}
\label{tab:response_quaility}
\end{table*}

\subsection{Results of Dialogue Policy}

\paragraph{Baselines.} We consider the following baselines:

\noindent (1) Direct: directly inference the LLM, with the same context.\\
\noindent (2) Retrieve: use RAG \cite{fan2024RAGmeetingLLMs} to retrieve the top-$2$ strategy. We employ E5-large \cite{wang2024E5} as the semantic retriever. \\
\noindent (3) Refine: a straightforward refinement method in which the model revises its initial response to incorporate emotional support considerations.\\
\noindent (4) Self-Refine: a method \citep{Madaan2023SelfRefine} initiated by generating feedback emphasizing emotional support from the initial response, then refining the response based on this feedback.\\
\noindent (5) CoT: uses the Chain-To-Thought prompt \citep{wei2022chain}, which first generates the seeker's \textit{emotion}, which then guides the generation of strategy and response.\\
\noindent (6) FSM: the finite state machine \cite{wangFSMFiniteState2024} with finite sets of states and state transitions triggered by inputs, and associated discrete actions.

\paragraph{Results.} Table \ref{tab:ID_results} shows the ID results of our dialogue policy $\pi(o)$, on the classification of emotion and strategy, as well as metrics of response. For most prompt-based baselines, it is difficult to classify the user emotion without pretrained knowledge, therefore we do not list this part of results. The only exception is FSM, which provides a detailed, situational strategy for the model to inference the emotion and strategy from finite sets. On the other hand, the finetuning-based baselines can classify both user emotion and the assistant strategy, with the training datasets organized accordingly. Nevertheless, our {\ModelName} consistently outperforms these baselines, on both emotion, strategy and response. Note we consider both similarity-based metrics (B-2 and R-L) and diversity-based metrics (D-2) here, which indicates a reasonable balance achieved by {\ModelName}. Table \ref{tab:detail_b_e} and \ref{tab:detail_b_q} in the Appendix further show per-emotion and per-strategy results, indicating {\ModelName} behaves equally across different emotions and strategies.

Table \ref{tab:OOD_results} further shows the OOD results on esconv and empathetic dialogues, from models trained by DailyDialog, conversations of daily topics. In this situation, {\ModelName} still generally performs better than baselines, with seldom exceptions. This observation ensures that the knowledge learned from general dialogues can smoothly transfer to some specific domains, \textit{i.e.}, the emphatic dialogues, with the assistance of the dialogue world model, as well as model-based reinforcement learning.

\paragraph{Human evaluations.} Table \ref{tab:response_quaility} presents human evaluation results. {\ModelName} achieves the highest scores in almost all aspects, which verifies the automatic evaluation results.


\paragraph{Cases.} Table \ref{tab:case_pi} showcases the dialogue policy and the dialogue world model, after being jointly trained by Algorithm 1. From this typical case, it can be observed that both $\pi$ and $q$ adapt to the specific domain well, provide accurate emotion and sentiment identification, and finally generate reasonable strategies and responses.

\begin{table*}[ht!]
\centering
\small
\resizebox{0.99\textwidth}{!}{
\begin{tabular}{l c c c p{10cm}}
      \toprule[1.5pt]
        \textbf{Roles} & \textbf{User's Emotion} & \textbf{User's Sentiment} & \textbf{Strategy} & \textbf{Response} \\ 
        \midrule
        \textit{User}& \textit{no emotion} & - & - &\textit{Hi there, I want to reserve a hotel room.} \\

        \textit{Assistant}& - & - & \textit{question} &\textit{No problem at all. Could I have your full name, please?} \\
        
        \textit{User}& \textit{no emotion} & - & - &\textit{Sure, John Sandals.} \\

        \textit{Assistant}& - & - & \textit{question} &\textit{Hi, Mr. Sandals. I'm Michelle, at your service. When do you need the room?} \\

        \textit{User}&\textit{no emotion} & - & - &\textit{My plans are to be there April 14th to the 17th.} \\
        
        \textit{Assistant}& - & - & \textit{question} &\textit{We have new room rates, sir. Will that be acceptable to you?} \\
        
        \midrule
        
        \textit{User}& \textit{surprise} & - & - & \textit{Well, it depends on the price, of course. What is it ?}\\
        
        \midrule
        \textit{{Assistant}}& - & - & \textit{{inform}} & {\textit{It's \$308 a night.}}\\
        Direct & no emotion & 0.5 & - & Our nightly rate currently sits at \$375. If you’d like, I can provide details about available amenities as well. \\
        \textbf{\ModelName} & {\textbf{surprise}} & {\textbf{0.43}} & {\textbf{inform}} & \textbf{The price is \$458 per night.} \\

        \midrule
        \textit{User}& \textit{no emotion} & - & - & \textit{I have no problem with that.}\\

        \midrule
        {\textit{Assistant}}& - & - & {\textit{question}} & {\textit{Great! Would you prefer smoking or nonsmoking?}}\\
        Direct & no emotion & 0.5 & - & Perfect, I’ll make a note of that. By the way, do you have any special requests or preferences during your stay? \\
        \textbf{\ModelName} & \textbf{happiness} & \textbf{0.78} & \textbf{question} & {\textbf{Would you like a smoking or non-smoking room?}} \\

        \midrule
        \textit{User}& \textit{disgust} & - & - & \textit{Definitely nonsmoking. I can't handle that smell.}\\

        \midrule
        {\textit{Assistant}} & - & - & {\textit{question}} & {\textit{Nonsmoking. Now, is a queen-size bed okay?}}\\
        Direct & disgust & 0.6 & - & Thanks for letting me know. Before we proceed, could you please confirm your email address? \\
        \textbf{\ModelName} & \textbf{{disgust}} & {\textbf{0.19}} & {\textbf{directive}} & {\textbf{Nonsmoking, nonsmoking. Now, if you'll just give me your phone number.}}\\

        \bottomrule
\end{tabular}
}
\caption{Typical cases generated by \ModelName  \ in DailyDialog. Contents from the original dataset are \textit{italic}, and results of {\ModelName} are \textbf{bolded}. Note that predictions of Emotion and Sentiment are for users instead of assistants.}
\label{tab:case_pi}
\end{table*}

\begin{table*}[htbp!]
\centering
\small
\begin{tabular}{l | ccc | ccc | ccc}
    \toprule
    \multicolumn{1}{c|}{\multirow{2}[2]{*}{Method}} &  \multicolumn{3}{c|}{Emotion} &  \multicolumn{3}{c|}{Strategy} &  \multicolumn{3}{c}{Response} \\ 
    \cmidrule{2-4} \cmidrule{5-7} \cmidrule{8-10}
    & ACC & MaF1 & $bias \downarrow$ & ACC & MaF1 & $bias \downarrow$ & B-2 & R-L & D-2 \\ 
    \toprule
    w/o WB & 87.67 & 43.36 & \underline{0.94} & 62.13 & 53.53 & 0.79 & 4.96 & 17.93 & 42.57 \\    
    w/o RL & 80.31 & 23.75 & 0.78 & 63.61 & \underline{56.87} & \underline{0.51} & 5.13 & 18.27 & 42.54 \\    
    w/o $b$ in $p$ & 86.71 & 41.36 & 1.19 & 61.13 & 52.68 & {0.54} & 6.16 & 19.26 & 42.75 \\
    w/o $b$ in $\mathcal{R}$ & \underline{87.86} & \underline{48.43} & \underline{0.94} & \underline{64.09} & 55.19 & 1.03 & \underline{11.04} & \underline{28.64} & \bf 49.55 \\
    single-model & 86.79 & 38.03 & 1.45 & 58.26 & 45.02 & 0.86 & 4.87 & 17.74 & 41.04 \\
    \bf {\ModelName} (ours) & \bf 88.05 & \bf 50.88 & \bf 0.74 & \bf 67.80 & \bf 62.29 & \bf 0.33 & \bf 11.65 & \bf 29.09 & \underline{49.36} \\
    \bottomrule
\end{tabular}
\caption{Ablation study on DailyDialog. The best results of each LLM are \textbf{bolded} and the second best are \underline{underlined}.}
\label{tab:ablation}
\end{table*}

\subsection{Ablation} To verify the effectiveness of the components of {\ModelName}, here we consider the following ablation settings:\\
\noindent $\bullet$ w/o WB: train the PPO policy without the knowledge of DWM.\\
\noindent $\bullet$ w/o RL: inference DWM directly, without the RL training.\\
\noindent $\bullet$ w/o $b$ in DWM: do not consider the user belief in the dialogue world model, \textit{i.e.}, only use dialogue history to predict the next-query of user.\\ 
\noindent $\bullet$ w/o $b$ in $\mathcal{R}$: do not consider the user belief in the reward model, \textit{i.e.}, provide the reward score based on the dialogue context only.\\ 
\noindent $\bullet$ single-model: make the policy and DWM a single parameter-shared model. 

As shown in Table \ref{tab:ablation}, {\ModelName} still performs the best on all the metrics, suggesting all its components are necessary to reach optimal performance. Especially, {\ModelName} utilizes the user belief information ($b$), resulting in further performance benefit compared to w/o $b$ in DWM and $\mathcal{R}$. Nevertheless, both w/o $b$ in DWM and w/o $b$ in $\mathcal{R}$ can still surpass the baselines in Table \ref{tab:ID_results}, indicating the pure application of MBRL on dialogue systems can substantially improve the performance. Last, the single-model approach can not behave as good as {\ModelName}, which indicates that it is still better to use separate models for the dialogue policy and the world model, given the current setting. 



\section{Related Work}


\paragraph{RL on dialogue system.}
RL has been widely applied to LLM-based dialogue systems by aligning models with human feedback via PPO \cite{ouyangTrainingLanguageModels2022a}. Further attempts like Q-star \cite{wangImprovingMultistepReasoning2024b} and ArCHer \cite{zhouArCHerTrainingLanguage2024b} improve the multi-step planning by value-based learning and hierarchical RL, respectively. To improve the sampling efficiency of traditional model-free RLs, there have also been applications of model-based RL (MBRL) on dialogue systems, such as DDQ \cite{peng-etal-2018-deep} and MCA \cite{xu-etal-2025-efficient}. Different from them, {\ModelName} includes the user belief in the LLM-based models, and solves a POMDP \citep{6407655}, enriching the exploration of dialogue policy by conditioning on model knowledge of user beliefs. 







\paragraph{World Models.} World Models \cite{haWorldModels2018} study the world dynamics primarily on vision-based inputs, such as PlaNet \cite{pmlr-v97-hafner19a}, Dreamer \cite{Hafner2020Dream} and Dream to Drive \cite{10547289}. There have also been world models on textual environments \cite{wu-etal-2021-gaussian, xu-etal-2025-efficient}. However, they are focused on task-oriented dialogues, while neglecting emotional cognition. On the contrary, this work proposes a dialogue world model for open-ended dialogues by explicitly modeling user beliefs.





\section{Conclusion}

In this paper, we propose a framework called {\ModelName} to introduce the MBRL on the dialogue system, with user belief modeling of emotion, sentiment and intention. We first pretrain a dialogue world model which allows the user emotional identification and the next-query prediction, then jointly train this world model with dialogue policy, to achieve better performance on the daily dialogues. We further verify the effectiveness of user belief both in the world model and the reward model, as well as the typical conversation cases.

\clearpage
\newpage

\section{Limitation}

Due to time and page limits, here we only explore a limited subset of user beliefs, including emotion, sentiment, and intention. Nevertheless, user belief modeling has the potential to consider more features, for example, user preferences, habits, and memory. A more thorough user modeling might further enhance the performance.

In addition to dialogue, language tasks have versatile scenarios, including question-answering, translation, summarization, and textual games. We expect this study could be a starting point for the world model application in textual environments, which may step forward in generalist artificial intelligence.

\section{Ethical Considerations}

{\ModelName} models the user belief, which might be correlated with the user's private information. Therefore, the confidentiality of datasets needs to be strictly confirmed. Also, by exposing the user's privacy on the screen, {\ModelName} can also potentially result in user inconvenience. Users should be aware of such conditions before deploying {\ModelName} on industrial applications.

\bibliography{custom}

\appendix

\section{More Implementation Details}
\label{sec:more_implement}

\subsection{Prompts} 
\label{sec:prompts}

Here we exhibit several prompts used in our framework, including the cognitive and generative parts of DWM prompts, and prompts of Actor, Critic and RM.

\paragraph{Prompt of DWM ($q(b_t|o_t)$).} The following prompt is utilized by the DWM model for emotion inference tasks. 




    \begin{tabular}{p{0.9\linewidth}}
\toprule
$prompt_{cognitive}$:\\
\hline
Below is a dialogue between a user and an assistant. The dialogue history is enclosed within <history> tags.\\

<history>  
\{history\}  
</history>\\

The user's current emotion before the assistant's last reply is: \{emotion\}.\\
The assistant's reply, employing the \{strategy\} strategy, is:  
\{assistant reply\}\\

Your task is to analyze the user's mental belief **after** receiving the assistant's reply. Complete the following three tasks based on the updated user emotion:

1. \textbf{Sentiment classification:}  
Classify the user's emotional polarity as either:  
-1 = negative, 0 = neutral, 1 = positive.  
Output format: \{"sentiment\_class": int\}

2. \textbf{Sentiment intensity regression:}  
Estimate the user's overall sentiment as a real number between 0 (extremely negative) and 1 (extremely positive).  
Output format: \{"sentiment\_score": float\}

3. \textbf{Emotion classification:}  
Classify the user's emotion into one or more of the following categories:  
\{no emotion, happiness, surprise, fear,  disgust, sadness, anger\}.  
Output format: \{"emotions": ["emotion1", "emotion2", ...]\}\\
 \hline
\end{tabular}

\paragraph{Prompt of DWM ($p(s_{t+1}|b_t, o_t)$).} The following prompt is utilized by the DWM model for next-query prediction. 

\begin{tabular}{p{0.9\linewidth}}
\toprule
$prompt_{generative}$: \\
\hline
Below is a dialogue between a user and an assistant. The dialogue history is enclosed within <history> tags. \\

<history>\\
\{history\}\\
</history> \\

The user's current emotion before the assistant's last reply is: \{emotion\}. \\
The assistant's reply, employing the \{strategy\} strategy, is: \\
\{assistant reply\} \\

If you are the user: \\
\textbf{1. Give the user's response after receiving this reply:} \\
\{user response\} \\
\vspace{1em}
Based on the updated user emotion after receiving the assistant's reply, complete the following tasks:

\textbf{2. Sentiment classification:} \\
Classify the user's emotional polarity as either: \\
\text{-1 = negative,\ \ 0 = neutral,\ \ 1 = positive} \\
Output format: \text{\{"sentiment\_class": int\}} \\

\textbf{3. Sentiment intensity regression:} \\
Estimate the user's overall sentiment as a real number between 0 (extremely negative) and 1 (extremely positive). \\
Output format: \text{\{"sentiment\_score": float\}} \\

\textbf{4. Emotion classification:} \\
Classify the user's emotion into one or more of the following categories:
\{no emotion, happiness, surprise, fear, disgust, sadness, anger\}
Output format: \text{\{"emotions": ["emotion1", "emotion2", ...]\}} \\

\bottomrule
\end{tabular}

\paragraph{Prompts of Actor, Critic and RM.} This prompt guides the assistant to first infer an appropriate conversational strategy based on the user's emotional state and dialogue history, and then generate a fitting response that aligns with that strategy.

The Critic and Reward model's prompt should be aligned with the Actor's to accurately evaluate the state value and reward.

\begin{table}[]
    \centering
    \begin{tabular}{p{0.9\linewidth}}
\toprule
$prompt_{RL}$:\\
\hline
Below is a dialogue between a user and an assistant. The dialogue history is enclosed within <history> tags.\\
<history>
\{history\}
</history>\\

User's emotion: \{belief\}

Given the user's emotion and the dialogue so far, first infer the most appropriate assistant strategy to move the dialogue forward.

Then, using the inferred strategy, the user's emotion, and the dialogue history, generate the next assistant response that naturally continues the dialogue.

Please output in the following format:

Assistant's strategy: \{strategy\}

Assistant's response: \{response\} \\
\hline
\end{tabular}
    \label{tab:my_label}
\end{table}


\subsection{Details of Datasets} 

Table \ref{tab:emotion_and_strategies} presents a comparison of three widely used emotion-centric dialogue datasets: ESConv, DailyDialog, and EmpatheticDialogues. Each dataset is annotated with both emotional categories and communication strategies (where available). ESConv includes a rich set of eight emotions and a diverse set of support strategies, which are abbreviated in the table for brevity. DailyDialog provides a smaller set of emotions along with basic dialogue act types. EmpatheticDialogues focuses primarily on emotional labels, covering a broader spectrum of feelings, with only the top 10 most frequent emotions shown here. This comparison highlights the varying granularity and scope of annotations across datasets used in empathetic and emotional dialogue research.

\begin{table*}[htbp!]
\centering
\small
\setlength{\tabcolsep}{6pt} 
\begin{tabular*}{\textwidth}{@{\extracolsep{\fill}} c | c | l }
\toprule
\multicolumn{1}{c|}{\textbf{Dataset}} & \multicolumn{1}{c|}{\textbf{Annotations}} & \multicolumn{1}{c}{\textbf{Types}}  \\ 
\toprule
\multirow{2}[0]{*}{ESconv} & Emotion & anger, anxiety, depression, disgust, fear, nervousness, sadness, shame \\ 
 & Strategy & Que., Paraphrasing \&Res., Ref., Self-Dis., Aff.\& Rea., Pro., Inf., Others \\ 
\midrule
\multirow{2}[0]{*}{DailyDialog} & Emotion & anger, disgust, fear, happiness, sadness, surprise, no emotion \\ 
 & Strategy & inform, question, directive, and commissive \\ 
\midrule
EmpatheticDialogues & Emotion & surprised, grateful, proud, sentimental, excited, sad, disgusted, angry, joyful, $\dots$ \\ 
\bottomrule
\end{tabular*}
\caption{Lists of emotions and strategies of ESConv, DailyDialog and EmpatheticDialogues. Strategies of ESconv here are abbreviated names; for full names, refer to the Appendix. Only the most frequent 9 emotions of EmpatheticDialogues are listed.}
\label{tab:emotion_and_strategies}
\end{table*}

Table \ref{tab:ex_ESConv} shows an example dialogue snippet from the ESConv dataset. It illustrates a conversation where the seeker expresses anxiety about quitting a disliked job without a secure alternative. The dialogue is annotated with the topic, the seeker’s query, the emotional state (anxiety with high intensity), and the empathetic strategy used by the supporter—in this case, a “reflection of feelings.” This example highlights how ESConv captures nuanced emotional expression alongside supportive conversational strategies.

\begin{table*}[htbp!]

    \centering
    \small
    
    \begin{tabular}{c|l}
        \toprule
        \textit{\textcolor{green}{Topic}} & \makecell[l]{\textcolor{green}{I hate my job but I am scared to quit and seek a} \textcolor{green}{new career.}} \\
        \midrule
        \textit{Query} & \makecell[l]{\textit{\{history\}} \\
        \textit{seeker:} Seriously!\\ What I'm scare of now is how to secure another job.}  \\
\midrule
        \textit{\textcolor{red}{Emotion}} & {\textcolor{red}{Anxiety} (intensity: 5)} \\
        \midrule
       \textit{\textcolor{blue}{Strategy}} & \makecell[l]{\textcolor{blue}{Reflection of feelings}}   \\
\midrule
      \textit{Response} & \makecell[l]{\textit{supporter:} I can feel your pain just by chatting with you.}   \\
        \bottomrule
    \end{tabular}
    \caption{An example of \textit{ESconv}.} 
    \label{tab:ex_ESConv}
\end{table*}

Table \ref{tab:statistics} presents a comparison of key statistics across three dialogue datasets: ESConv, DailyDialog, and EmpatheticDialogues. It includes data on the number of sessions, utterances, average utterance lengths, and speaker-specific information such as utterance counts, average lengths, and the number of annotated strategies and emotions.

\begin{table*}[ht!] 
  \centering
  \resizebox{0.85\textwidth}{!}{
    \begin{tabular}{llccc}
    \toprule
    \multicolumn{2}{c}{Category ($\downarrow$)} & ESconv & DailyDialog & EmpatheticDialogues (test set only)   \\
    \midrule
    \multicolumn{2}{l}{\# Sessions} & 1.3K & 13.1k& 2.5K \\
    \multicolumn{2}{l}{\# Utterances} & 38K & 103.0k& 11.0K\\
    \multicolumn{2}{l}{Average \# Utterances} & 28.9  & 7.9& 4.3\\
    \multicolumn{2}{l}{Average Utterance Length} & 18.8  & 13.6& 16.7\\
    \midrule
    \multirow{5}[0]{*}{Seeker/Speaker1} & \# Utterances & 20K&53.8k& 5.7K \\
       & Avg \# Utterances & 15.4 & 4.1& 2.2 \\
       & Avg Uttr Len & 16.8& 13.2& 20.8  \\
        & \# Strategies & -& 4& -\\
       & \# Emotions & 11& 7& 32 \\
    \midrule
    \multirow{5}[0]{*}{Supporter/Speaker2} & \# Utterances & 18K & 49.2k& 5.2K\\
       & Avg \# Utterances & 13.6& 3.9& 2.1  \\
       & Avg Uttr Len & 21.0& 14.1& 12.3  \\
       & \# Strategies & 8& 4& -\\
       & \# Emotions & -& 7& 32 \\
    \bottomrule
    \end{tabular}%
  }
  \caption{
  Statistics of ESConv, DailyDialog and EmpatheticDialogues. 
  }
  \label{tab:statistics}%
\end{table*}

\subsection{Metrics of Classification and Regression}


\paragraph{F1-scores.} F1-related scores include Micro-F1 and Macro-F1. Micro-F1 considers the overall precision and recall of all instances, while Macro-F1 equals the average F1-score of labels.

\paragraph{$bias$.} We define the preference $bias$ as \textit{how much the model prefers certain labels over others}. To quantify the preference for each strategy in LLMs, we employ the Bradley-Terry model~\citep{bradley1952btmodel}, which is widely used in human preference modeling~\citep{Rafailov2023DPOLMisReward}. Following~\citet{Newman2023Efficient_BT}, we formally derive the preference $p$ for strategy $i$ as follows:
\begin{equation}
\normalsize
    p_{i}' =  \frac{\sum_{j}(w_{ij}p_{j})/(p_{i}+p_{j})}{\sum_{j}w_{ji}/(p_{i}+p_{j})} 
\label{eq:BT_equation}
\end{equation}
where $w_{ij}$ represents the number of times the model predicts strategy $i$ when the ground-truth strategy is $j$.
All of the preferences $p_i$ are initialized as 1 and updated through iteration of the Eq~(\ref{eq:BT_equation})
, where $p_i'$ represents the preference in the next iteration.
After the final iteration, we scale the total sum of $p_i$ to 8 ($\sum{p_i}=8$) so that the average $\bar{p}$ becomes 1, indicating a strong preference for strategy $i$ if $p_i>1$.

We use a standard deviation of preferences $p_i$ across the strategies as $bias$.
\begin{equation}
\normalsize
    bias = \sqrt{\frac{\sum_{i=1}^{N}(p_i - \bar{p})^2}{N}}
\end{equation}
where a higher value for $bias$ indicates that the model exhibits a clear preference for both preferred and non-preferred strategies \citep{kang-etal-2024-large}.

\paragraph{Pearson Correlation Coefficient.}

The Pearson correlation coefficient $r$ provides a dimensionless index of the linear relationship between two continuous variables $x$ and $y$. Formally, $r$ is defined as

\begin{equation}
r = \frac{\displaystyle\sum_{i=1}^{n}(x_{i}-\bar{x})(y_{i}-\bar{y})}{\sqrt{\displaystyle\sum_{i=1}^{n}(x_{i}-\bar{x})^{2}}\;\sqrt{\displaystyle\sum_{i=1}^{n}(y_{i}-\bar{y})^{2}}}
\end{equation}

\subsection{Metrics of Generation}



\paragraph{BLEU-2.} B-2\citep{papineni2002bleu} first compute the geometric average of the modified $n$-gram precisions, $p_n$, using $n$-grams up to length $N$ and positive weights $w_n$ summing to one.

Next, let $c$ be the length of the prediction and $r$ be the reference length. The BP and BLEU-2 are computed as follows.

\begin{equation}
    \mathrm{BP}=\left\{\begin{array}{ll}
1 & \text { if } c>r \\
e^{(1-r / c)} & \text { if } c \leq r
\end{array} .\right.
\end{equation}

\begin{equation}
    \mathrm{BLEU}=\mathrm{BP} \cdot \exp \left(\sum_{n=1}^N w_n \log p_n\right) .
\end{equation}

\paragraph{Rouge-L.} R-L\citep{lin2004rouge} propose using LCS-based F-measure to estimate the similarity between two summaries $X$ of length $m$ and $Y$ of length $n$, assuming $X$ is a reference summary sentence and $Y$ is a candidate summary sentence, as follows:

\begin{equation}
\begin{aligned}
& R_{l c s}=\frac{L C S(X, Y)}{m} \\
& P_{l c s}=\frac{L C S(X, Y)}{n} \\
& F_{l c s}=\frac{\left(1+\beta^2\right) R_{l c s} P_{l c s}}{R_{l c s}+\beta^2 P_{l c s}}
\end{aligned}
\label{rouge_l}
\end{equation}

Where $\operatorname{LCS}(X, Y)$ is the length of a longest common subsequence of $X$ and $Y$, and $\beta=P_{l c s} / R_{\text {lcs }}$ when $\partial F_{l c s} / \partial R_{l c s}=\partial F_{l c s} / \partial P_{l c s}$. In DUC, $\beta$ is set to a very big number $(\rightarrow \infty)$. Therefore, the LCS-based F-measure, \textit{i.e.}, Equation \ref{rouge_l}, is Rouge-L. 

\paragraph{Dist-2.} \citet{li2015diversity} report the degree of diversity by calculating the number of distinct unigrams and bigrams in generated responses.
The value is scaled by the total number of generated tokens to avoid favoring long sentences:
\begin{equation} \label{eq:4}
Dist(n) = \frac{Count(unique\ n-gram)}{Count(n-gram)}
\end{equation}

\subsection{Principle of Human Scoring}
\label{sec:huam_score_principle}

We start with the criteria proposed by \citet{kang-etal-2024-large}. The human evaluation is aimed to align with the ultimate purpose of ESC, the seeker's \textit{satisfaction}. To achieve this, the supporter's behavior can be further classified into the following criteria:

\noindent \textit{Acceptance}: Does the seeker accept without discomfort;

\noindent \textit{Effectiveness}: Is it helpful in shifting negative emotions or attitudes towards a positive direction; 

\noindent \textit{Sensitivity}: Does it take into consideration the general state of the seeker. Furthermore, to clarify the capability of LLMs to align strategy and responses, we include Alignment.

To achieve a more elaborate assessment, we consider three more dimensions addressing the generation quality:

\noindent \textit{Fluency}: the level of fluency of response.

\noindent \textit{Emotion}: the emotional intensity of response which could affect the seeker's emotional state.

\noindent \textit{Interesting}: Whether the response can arouse the seeker's interest and curiosity, presenting unique ideas, vivid expressions or engaging elements that capture the seeker's attention and make the interaction more appealing.


We invited 10 interns as annotators for the human evaluation. From the test set, we sampled 10 dialogue sessions, and each annotator scored all responses independently. We then calculated the average score for each method, as shown in Table \ref{tab:response_quaility}. Regarding statistical significance, we performed a t-test on the average scores of different methods to verify whether our method significantly outperforms the others  in Table\ref{tab:response_significance}. 

The interns rate the models according to these multiple aspects, namely Fluency, Emotion, Interesting, and Satisfaction, with Satisfaction covering Acceptance, Effectiveness, Sensitivity, and Satisfaction itself. \\
Throughout this evaluation process, we strictly comply with international regulations and ethical norms, ensuring that all practices conform to the necessary guidelines regarding participant involvement and data integrity.\\
annotators are required to independently evaluate each sample in strict accordance with the pre-established criteria. By adhering to these principles, the evaluation process maintains objectivity, standardization, and consistency, thus enhancing the overall quality and credibility of the evaluation results. To ensure the reliability of the results, we also evaluated the consistency of the evaluations  in Table\ref{tab:evaluation_correlation} to ensure that the annotator did not score arbitrarily.
\\
The detailed manual scoring criteria are as follows:
\begin{itemize}
\item Fluency:

1: The sentence is highly incoherent, making it extremely difficult to understand and failing to convey a meaningful idea.

2: The sentence has significant incoherence issues, with only parts of it making sense and struggling to form a complete thought.

3: The sentence contains some incoherence and occasional errors, but can still convey the general meaning to a certain extent.

4: The sentence is mostly fluent with only minor errors or slight awkwardness in expression, and effectively communicates the intended meaning.

5: Perfect. The sentence is completely fluent, free of any errors in grammar, punctuation, or expression, and clearly conveys the idea.

\item Emotion:

1: The emotional expression is extremely inappropriate and chaotic, not in line with the content, and may convey wrong emotions.

2: The emotional expression has obvious flaws, either too weak or exaggerated, and is disjointed from the content.

3: The emotional expression is average. It can convey basic emotions but lacks depth and has minor issues.

4: The emotional expression is good. It can effectively convey the intended emotion with an appropriate intensity and is well integrated with the content.

5: The emotional expression is excellent. It is rich, nuanced, and perfectly matches the content, capable of evoking a strong and appropriate emotional response.

\item Acceptance:

1: The response inescapably triggers emotional resistance.

2: The response is highly likely to trigger emotional resistance.

3: The response has a possibility of emotional resistance occurring.

4: The response rarely provokes emotional resistance.

5: The response has no occurrence of emotional resistance.

\item Effectiveness:

1:  The response actually worsens the seeker's emotional distress.

2: The response carries the risk of increasing stress levels, and this outcome varies depending on the individual user.

3: The response fails to alter the seeker's current emotional intensity and keeps it at the same level.

4: The response shows promise in calming the emotional intensity; however, it is overly complicated or ambiguous for the user to fully comprehend and utilize effectively.

5: The response appears to be highly effective in soothing the seeker's emotions and offers valuable and practical emotional support. 

\item Sensitivity:

1: The response renders inaccurate evaluations regarding the seeker's state.

2: The response is characterized by rash judgments, as it lacks adequate assessment and in-depth exploration of the seeker's state.

3: The response is formulated with a one-sided judgment and a limited exploration of the seeker's state.

4: The response demonstrates an understanding that only covers a part of the seeker's state.

5: The response precisely grasps the seeker's state and is appropriately tailored according to the seeker's actual situation.

\item Alignment:

1: The response is in total contradiction to the predicted strategy.

2: The response has a minor deviation from the predicted strategy.

3: There is some ambiguity between the response and the predicted strategy.

4: The response largely matches the predicted strategy, yet it contains some ambiguous elements.

5: The response effectively makes itself consistent with the predicted strategy.

\item Satisfaction:

1: The response is extremely disappointing. It doesn't answer the question at all and is of no help.

2: The response is poor. It only gives a partial answer and leaves many doubts unresolved.

3: The response is average. It meets the basic requirements but isn't particularly outstanding.

4: The response is good. It answers the question clearly and provides some useful details.

5: The response is excellent. It not only answers the question perfectly but also offers valuable additional insights.

\end{itemize}

\section{More Results}

\subsection{Evidence Lower Bound Derivations and Discussion}
\label{sec:append-ELBO}

The variational bound for latent dynamics models
$p\left(o_{1: T}, b_{1: T} \mid a_{1: T}\right)=\prod_tp(b_t|b_{t-1},a_{t-1})p(o_t|b_t)$
 and a variational posterior $q\left(b_{1: T} \mid o_{1: T}, a_{1: T}\right)=\prod_{t} q\left(b_{t} \mid o_{\leq t}, a_{<t}\right)$ follows from importance weighting and Jensen's inequality as shown,

{\small
\begin{equation}
\begin{aligned}
&\log p\left(o_{1: T},r_{1: T}  | a_{1: T}\right) \\
&= 
\log \mathrm{E}_{p\left(b_{1: T} | a_{1: T}\right)}
\left[\prod_{t=1}^{T} p\left(o_{t} | b_{t}\right)\mathcal{R}\left(r_{t} | b_{t}\right)\right] \\
&= \log \mathrm{E}_{q\left(\textbf{b} | \textbf{o}, \textbf{a}\right)} 
\left[\prod_{t=1}^{T} \frac{p\left(o_{t} | b_{t}\right)
p\left(b_{t} | b_{t-1}, a_{t-1}\right)}
{q\left(b_{t} | o_{\leq t}, a_{<t}\right)}\mathcal{R}\left(r_{t} | b_{t}\right)\right] \\
&\geq \mathrm{E}_{q\left(b_{1: T} | o_{1: T}, a_{1: T}\right)}\left[
\sum_{t=1}^{T} \log p\left(b_{t} | b_{t-1}, a_{t-1}\right) \right. \\
&\left. \quad - \log q\left(b_{t} | o_{\leq t}, a_{<t}\right) + \log p\left(o_{t} | b_{t}\right)+\log\mathcal{R}\left(r_{t} | b_{t}\right)\right]
\end{aligned}
\end{equation}
}
, where $\textbf{b} = b_{1: T},\textbf{a} = a_{1: T},\textbf{o} = o_{1: T}$.

In our work, we choose ELBO as the optimization object in Eq.\ref{Eq:Jmodel} as we aim to maximize the log-likelihood 
, which represent our world dialogue model. Here, observation, reward, and action are all observable variables. We introduce the belief state to better model the user’s dialogue trajectory and optimize the belief to maximize the log-likelihood given the actions.

Since this likelihood is intractable, we derive a tractable lower bound using the classical ELBO formulation, resulting in the right-hand side of Eq.\ref{Eq:Jmodel}, with detailed derivation provided above. Therefore, we optimize this lower bound as a surrogate for the original log-likelihood to train our DWM.

To further explain our algorithm, we can divide Equation 4 into two terms.

The first term is the reconstruction term, which represents the expected log-likelihood of reconstructing the data (i.e., observations and rewards) under the posterior of the belief state. We aim to maximize this term.

The second term is the KL divergence, acting as a regularization term that penalizes the difference between the posterior 
 and prior 
 distributions over belief states. It encourages the posterior not to deviate excessively from the prior, which helps prevent overfitting and improves generalization.

\subsection{More result curves}

\begin{figure}[htbp]
    \centering
    \includegraphics[width=0.45\linewidth]{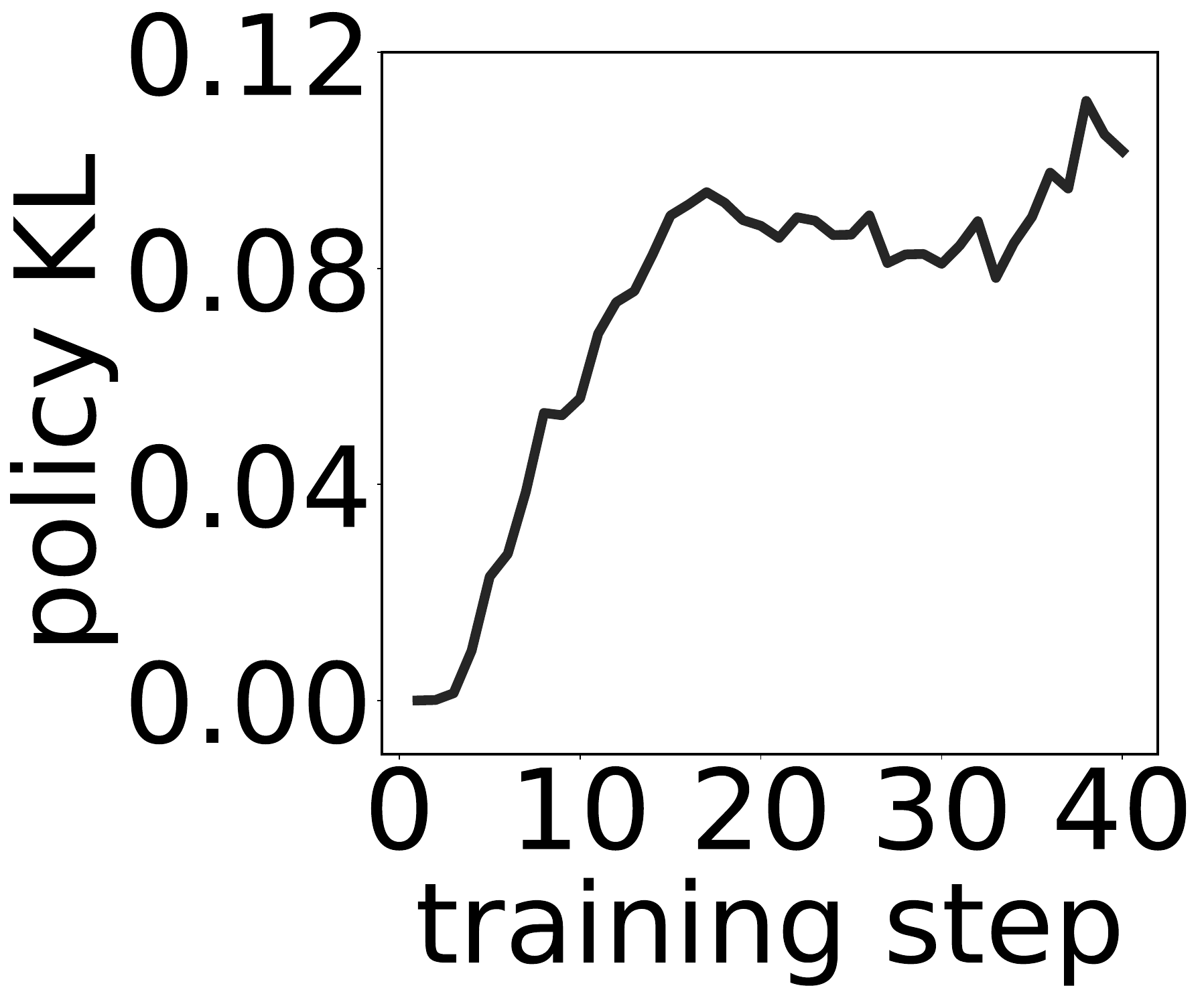}
    \hspace{0.1in}
    \includegraphics[width=0.45\linewidth]{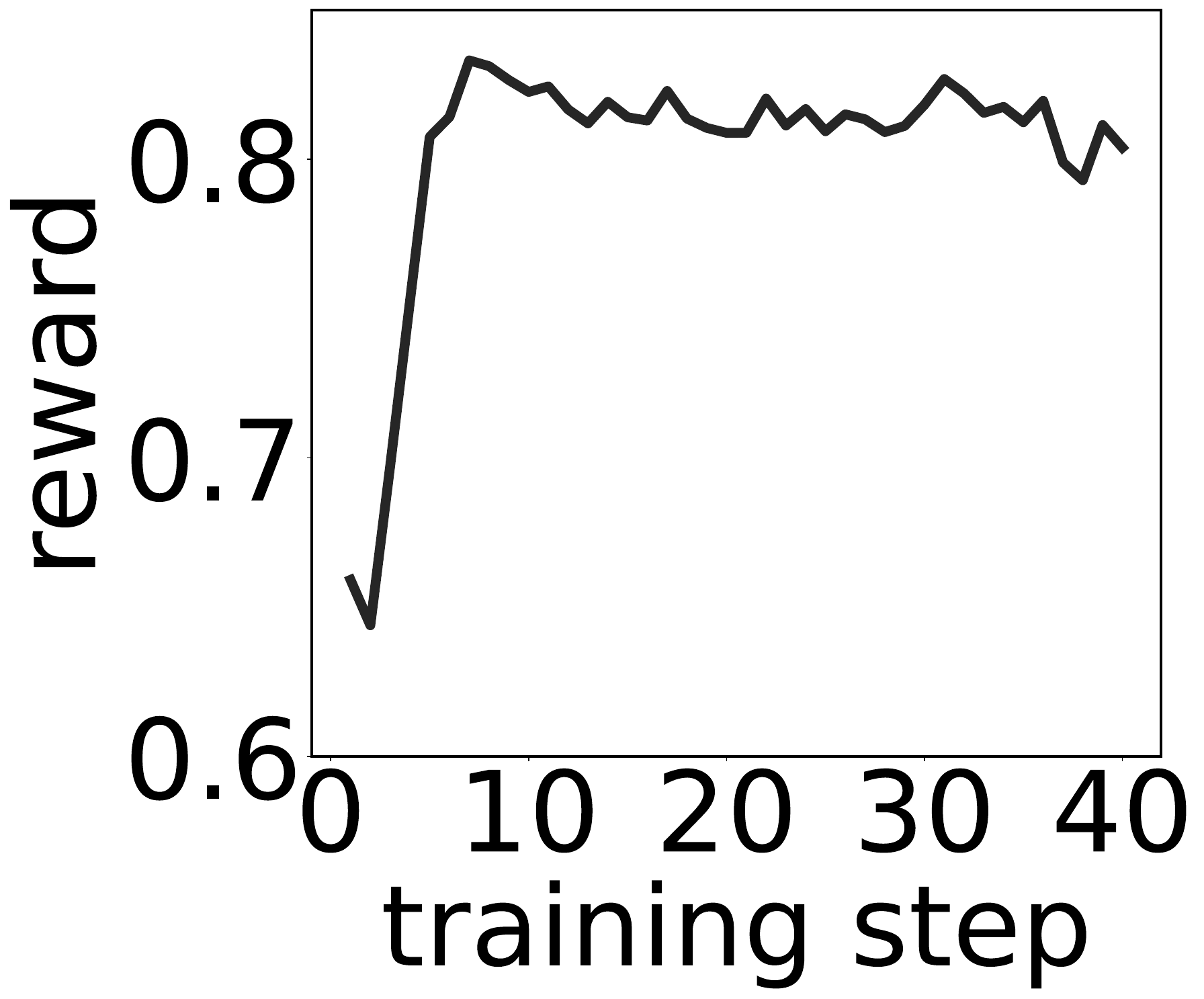}
    \caption{More training plots of {\ModelName}, including the policy KL (left) and reward (right).}
    \label{fig:more_curves}
\end{figure}

Figure \ref{fig:more_curves} shows the training dynamics of {\ModelName}. The left plot illustrates the policy KL divergence, which reflects the difference between the current policy and the reference model. While KL naturally increases during PPO training, we keep it within a controlled range to maintain stability. The right plot shows the reward steadily increasing and eventually converging, indicating good training stability and convergence.

\begin{figure}
    \centering
    \includegraphics[width=0.95\linewidth]{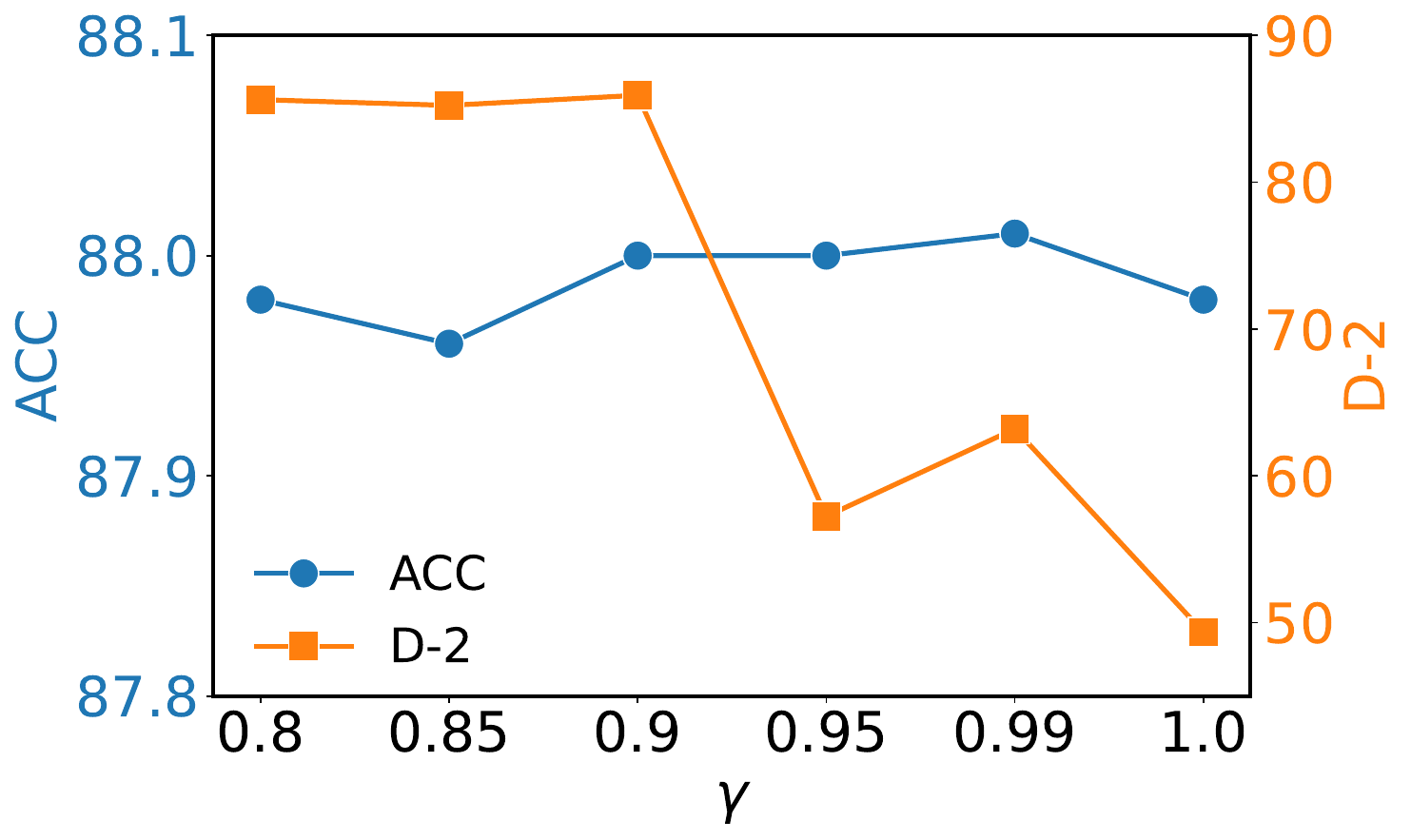}
    \caption{Curves of Acc and D-2 variations under different gamma values.}
    \label{fig:gamma}
\end{figure}

As shown in Figure \ref{fig:gamma}, although the Acc is slightly higher when gamma is set to 1.0, the D-2 metric drops significantly. Considering both indicators, setting gamma to 0.9 achieves the best overall performance and brings out the full potential of the algorithm.

\begin{table*}[t!]
\small
\centering
\begin{tabular}{c|l|cccccccc}
    \toprule
     & \multicolumn{1}{c|}{\multirow{2}[2]{*}{Model}} & \multicolumn{8}{c}{Emotion}   \\
    \cmidrule{3-10}
  &   & \textbf{no emo} & \textbf{happiness} & \textbf{surprise} & \textbf{fear} & \textbf{disgust} & \textbf{sadness} & \textbf{anger} & \textbf{total} \\
  
    \midrule
     \multicolumn{1}{c|}{\multirow{4}[2]{*}{ACC}}
&+ SFT & 91.65 & 0.00 & 23.00 & 0.00 & 2.63 & 0.00 & 0.00 & 76.76 \\
&+ COT+SFT & 99.10 & 8.09 & 1.00 & 0.00 & 0.00 & 0.00 & 1.14 & 83.48 \\
&+ FSM+SFT & 99.81 & 0.62 & 0.00 & 0.00 & 0.00 & 5.26 & 0.00 & 83.28 \\
&\textbf{{\ModelName}} & 95.65 & 56.61 & 55.00 & 21.43 & 15.79 & 31.58 & 32.95 & 88.05 \\

    \midrule
     \multicolumn{1}{c|}{\multirow{4}[2]{*}{MaF1}}
&+ SFT & 87.17 & 0.00 & 8.13 & 0.00 & 5.13 & 0.00 & 0.00 & 14.35 \\
&+ COT+SFT & 90.96 & 14.34 & 1.72 & 0.00 & 0.00 & 0.00 & 2.15 & 15.60 \\
&+ FSM+SFT & 90.89 & 1.23 & 0.00 & 0.00 & 0.00 & 8.99 & 0.00 & 14.44 \\
&\textbf{{\ModelName}} & 93.17 & 62.81 & 56.70 & 30.00 & 27.27 & 44.44 & 41.73 & 50.88 \\

\midrule
 \multicolumn{1}{c|}{\multirow{4}[2]{*}{$bias$}}
&+ SFT & 2.21 & 1.23 & 2.45 & 2.45 & 1.07 & 2.45 & 1.57 & 2.03 \\
&+ COT+SFT & 0.66 & 1.98 & 1.61 & 2.45 & 1.50 & 1.74 & 2.45 & 1.98 \\
&+ FSM+SFT & 0.78 & 1.99 & 2.45 & 2.45 & 2.45 & 2.45 & 1.79 & 2.22 \\
&\textbf{{\ModelName}} & 0.65 & 1.52 & 1.05 & 2.45 & 1.42 & 2.45 & 1.07 & 0.74\\
\bottomrule
\end{tabular}
\caption{Per-emotion automatic metrics on DailyDialog.}
\label{tab:detail_b_e}
\end{table*}

\subsection{Per-emotion automatic metrics}
\label{sec:per-emotion}

Table \ref{tab:detail_b_e} presents the performance of different models across four dialogue emotions. Notably, our model \ModelName demonstrates a more uniform distribution of performance across different emotional categories in various metrics, thereby mitigating emotion-related bias.

\begin{table}[h]
\centering
\resizebox{0.5\textwidth}{!}{
\begin{tabular}{c|l|ccccc}
    \toprule
     & \multicolumn{1}{c|}{\multirow{2}[2]{*}{Model}} & \multicolumn{5}{c}{Strategy}   \\
    \cmidrule{3-7}
  &   & \textbf{directive} & \textbf{inform} & \textbf{question} & \textbf{commissive} & \textbf{total}  \\ 
  
    \midrule
     \multicolumn{1}{c|}{\multirow{4}[2]{*}{ACC}}
&+ SFT & 1.30 & 78.85 & 47.00 & 74.77 & 60.19 \\
&+ COT+SFT & 0.37 & 78.02 & 51.88 & 69.91 & 60.11 \\
&+ FSM+SFT & 3.15 & 85.85 & 50.75 & 67.28 & 64.05 \\
&\textbf{{\ModelName}} & 42.79 & 80.83 & 58.41 & 68.34 & 67.80 \\

    \midrule
     \multicolumn{1}{c|}{\multirow{4}[2]{*}{MaF1}}
&+ SFT & 2.55 & 75.86 & 44.24 & 56.62 & 44.82 \\
&+ COT+SFT & 0.74 & 76.01 & 44.67 & 58.19 & 44.90 \\
&+ FSM+SFT & 6.01 & 78.48 & 49.78 & 59.17 & 48.36 \\
&\textbf{{\ModelName}} & 48.53 & 77.78 & 61.38 & 61.46 & 62.29 \\

\midrule
 \multicolumn{1}{c|}{\multirow{4}[2]{*}{$bias$}}
&+ SFT & 0.60 & 0.76 & 0.77 & 0.73 & 0.82 \\
&+ COT+SFT & 0.60 & 0.76 & 0.77 & 0.73 & 0.82 \\
&+ FSM+SFT & 0.61 & 0.83 & 0.77 & 0.77 & 0.66 \\
&\textbf{{\ModelName}} & 0.62 & 0.59 & 0.65 & 0.60 & 0.33 \\
  \midrule
 \multicolumn{1}{c|}{\multirow{4}[2]{*}{B-2}}
&+ SFT & 4.45 & 7.25 & 6.74 & 7.96 & 6.81 \\
&+ COT+SFT & 4.61 & 6.80 & 7.25 & 7.07 & 6.61 \\
&+ FSM+SFT & 6.50 & 5.50 & 7.05 & 4.44 & 5.85 \\
&\textbf{{\ModelName}} & 10.20 & 12.38 & 12.11 & 9.42 & 11.65 \\
  \midrule
  \multicolumn{1}{c|}{\multirow{4}[2]{*}{R-L}}
&+ SFT & 14.59 & 19.92 & 17.00 & 19.72 & 18.54 \\
&+ COT+SFT & 14.69 & 19.13 & 17.74 & 18.22 & 18.09 \\
&+ FSM+SFT & 21.28 & 21.50 & 23.02 & 21.20 & 21.80 \\
&\textbf{{\ModelName}} & 25.15 & 30.62 & 28.14 & 30.38 & 29.09 \\
\midrule
  \multicolumn{1}{c|}{\multirow{4}[2]{*}{D-2}}
&+ SFT & 59.82 & 53.18 & 55.81 & 58.77 & 43.36 \\
&+ COT+SFT & 58.03 & 53.18 & 54.25 & 56.37 & 42.87 \\
&+ FSM+SFT & 62.07 & 55.83 & 54.10 & 60.59 & 47.43 \\
&\textbf{{\ModelName}} & 66.25 & 59.24 & 59.15 & 67.77 & 49.36 \\
\bottomrule
\end{tabular}
}
\caption{Per-strategy metrics on DailyDialog.}
\label{tab:detail_b_q}
\end{table}

\subsection{Per-strategy automatic metrics}\label{sec:per-strategy}

Table \ref{tab:detail_b_q} presents the performance of different models across four dialogue emotions on the DailyDialog dataset, using several automatic evaluation metrics. Overall, {\ModelName} consistently outperforms the baselines across all metrics, demonstrating stronger generation quality and better strategic alignment.





\subsection{Significance Test for Human Evaluation}

To verify the reliability of manual scoring, we performed the following T-test on the results based on the means and standard deviations in Table \ref{tab:response_quaility}.

Table \ref{tab:response_significance} presents our additional significance testing results, which report the p-value of the following hypothesis test:

\begin{equation}\label{hypothesis_test}
    H_0: Metric_X > Metric_DreamCUB
\end{equation}

The results indicate that our method achieves statistically significant improvements over most of the baselines.

\begin{table*}[htbp!]
\centering
\small
\resizebox{0.98\textwidth}{!}{%
\begin{tabular}{l|ccccccc}
    \toprule
 \multicolumn{1}{c|}{Method} & Fluency  & Emotion & Acceptance & Effectiveness & Sensitivity & Alignment & Satisfaction \\ 
    \toprule   
    Direct & <0.01 & <0.01 & <0.01 & <0.01 & <0.01 & <0.01 & <0.01  \\  
    \;+Refine & <0.01 & <0.05 & <0.01 & <0.01 & <0.01 & <0.01 & <0.01  \\  
    \;+Self-Refine & <0.01 & <0.05 & <0.01 & <0.01 & <0.01 & <0.01 & <0.01  \\ 
    \;+CoT & <0.01 & <0.05 & <0.01 & <0.01 & <0.01 & <0.01 & <0.01  \\ 
    \;+FSM & <0.01 & 0.2 & <0.01 & <0.01 & <0.01 & <0.01 & <0.01  \\ 
    \midrule
    \;+SFT & <0.01 & 0.27 & <0.01 & <0.01 & <0.01 & <0.01 & <0.01  \\ 
    \;+CoT+SFT & 0.14 & 0.68 & <0.01 & <0.01 & <0.05 & <0.01 & <0.01  \\  
    \;+FSM+SFT & 0.38 & 0.56 & <0.01 & 0.09 & 0.35 & <0.05 & <0.05  \\  
    \;+DreamCUB & -- & -- & -- & -- & -- & -- & --  \\ 
    \bottomrule
    \end{tabular}%
}
\caption{P-value of the hypothesis test on human evaluation.}
\label{tab:response_significance}
\end{table*}

\begin{table*}[!ht]
    \centering
    \small
    \begin{tabular}{ccccccccccc}
    \toprule 
    ~ & $0$ & $1$ & $2$ & $3$ & $4$ & $5$ & $6$ & $7$ & $8$ & $9$  \\ 
    \midrule
    $0$ & 1 & 0.71 & 0.46 & 0.41 & 0.58 & 0.67 & 0.6 & 0.61 & 0.49 & 0.58  \\ 
    $1$ & 0.71 & 1 & 0.66 & 0.46 & 0.65 & 0.7 & 0.59 & 0.56 & 0.61 & 0.63  \\ 
    $2$ & 0.46 & 0.66 & 1 & 0.34 & 0.49 & 0.64 & 0.5 & 0.44 & 0.59 & 0.68  \\ 
    $3$ & 0.41 & 0.46 & 0.34 & 1 & 0.39 & 0.39 & 0.46 & 0.37 & 0.46 & 0.43  \\ 
    $4$ & 0.58 & 0.65 & 0.49 & 0.39 & 1 & 0.55 & 0.54 & 0.59 & 0.53 & 0.56  \\ 
    $5$ & 0.67 & 0.7 & 0.64 & 0.39 & 0.55 & 1 & 0.62 & 0.52 & 0.56 & 0.69  \\ 
    $6 $& 0.6 & 0.59 & 0.5 & 0.46 & 0.54 & 0.62 & 1 & 0.54 & 0.57 & 0.59  \\ 
    $7$ & 0.61 & 0.56 & 0.44 & 0.37 & 0.59 & 0.52 & 0.54 & 1 & 0.58 & 0.61  \\ 
    $8$ & 0.49 & 0.61 & 0.59 & 0.46 & 0.53 & 0.56 & 0.57 & 0.58 & 1 & 0.61  \\ 
    $9$ & 0.58 & 0.63 & 0.68 & 0.43 & 0.56 & 0.69 & 0.59 & 0.61 & 0.61 & 1 \\ 
    \bottomrule
    \end{tabular}
\caption{The correlation matrix of human evaluations on fluency. Row $i$ and Column $j$ denote the $i$ and $j$-th annotators, respectively.}
\label{tab:evaluation_correlation}
\end{table*}

\subsection{Consistency of human evaluation}

To assess the consistency of annotators' scores, we computed the correlation matrix of scores across all 10 annotators for each evaluation dimension. The correlation matrix of 'fluency' is presented in Table \ref{tab:evaluation_correlation}. Except for Annotator 3, the others maintained relatively high levels of inter-agreement. Results of other dimensions are similar, and we omit them for ease of clarity.

\end{document}

%% file: math_commands.tex
\usepackage{amsmath,amsthm,amsfonts,bm,graphicx,tikz,wrapfig,algorithm,algorithmicx,algpseudocode,listings,tabularx}
\tikzstyle{arr} = [-stealth, thick]

\newtheorem{theorem}{Theorem}

\def\eqref#1{equation~\ref{#1}}

\def\1{\bm{1}}

\def\vx{{\bm{x}}}
\def\vy{{\bm{y}}}

\DeclareMathAlphabet{\mathsfit}{\encodingdefault}{\sfdefault}{m}{sl}
\SetMathAlphabet{\mathsfit}{bold}{\encodingdefault}{\sfdefault}{bx}{n}

\newcommand{\KL}{D_{\mathrm{KL}}}



%% file: main.bbl
\begin{thebibliography}{43}
\providecommand{\natexlab}[1]{#1}

\bibitem[{AI@Meta(2024)}]{llama3modelcard}
AI@Meta. 2024.
\newblock \href {https://github.com/meta-llama/llama3/blob/main/MODEL_CARD.md}
  {Llama 3 model card}.

\bibitem[{Bradley and Terry(1952)}]{bradley1952btmodel}
Ralph~Allan Bradley and Milton~E Terry. 1952.
\newblock Rank analysis of incomplete block designs: I. the method of paired
  comparisons.
\newblock \emph{Biometrika}, 39(3/4):324--345.

\bibitem[{Chen et~al.(2022)Chen, Mu, Luo, Li, and Chen}]{pmlr-v162-chen22q}
Xiaoyu Chen, Yao~Mark Mu, Ping Luo, Shengbo Li, and Jianyu Chen. 2022.
\newblock \href {https://proceedings.mlr.press/v162/chen22q.html} {Flow-based
  recurrent belief state learning for {POMDP}s}.
\newblock In \emph{Proceedings of the 39th International Conference on Machine
  Learning}, volume 162 of \emph{Proceedings of Machine Learning Research},
  pages 3444--3468. PMLR.

\bibitem[{Deisenroth and
  Rasmussen(2011)}]{deisenrothPILCOModelbasedDataefficient2011}
Marc~Peter Deisenroth and Carl~Edward Rasmussen. 2011.
\newblock {{PILCO}}: A model-based and data-efficient approach to policy
  search.
\newblock In \emph{Proceedings of the 28th {{International Conference}} on
  {{International Conference}} on {{Machine Learning}}}, {{ICML}}'11, pages
  465--472, Madison, WI, USA. Omnipress.

\bibitem[{Demszky et~al.(2020)Demszky, Movshovitz-Attias, Ko, Cowen, Nemade,
  and Ravi}]{demszky-etal-2020-goemotions}
Dorottya Demszky, Dana Movshovitz-Attias, Jeongwoo Ko, Alan Cowen, Gaurav
  Nemade, and Sujith Ravi. 2020.
\newblock \href {https://doi.org/10.18653/v1/2020.acl-main.372}
  {{G}o{E}motions: A dataset of fine-grained emotions}.
\newblock In \emph{Proceedings of the 58th Annual Meeting of the Association
  for Computational Linguistics}, pages 4040--4054, Online. Association for
  Computational Linguistics.

\bibitem[{Fan et~al.(2024)Fan, Ding, Ning, Wang, Li, Yin, Chua, and
  Li}]{fan2024RAGmeetingLLMs}
Wenqi Fan, Yujuan Ding, Liangbo Ning, Shijie Wang, Hengyun Li, Dawei Yin,
  Tat-Seng Chua, and Qing Li. 2024.
\newblock \href {https://doi.org/10.1145/3637528.3671470} {A survey on rag
  meeting llms: Towards retrieval-augmented large language models}.
\newblock In \emph{Proceedings of the 30th ACM SIGKDD Conference on Knowledge
  Discovery and Data Mining}, KDD '24, page 6491–6501, New York, NY, USA.
  Association for Computing Machinery.

\bibitem[{Firdaus et~al.(2023)Firdaus, Singh, Ekbal, and
  Bhattacharyya}]{10.1145/3583780.3615265}
Mauzama Firdaus, Gopendra Singh, Asif Ekbal, and Pushpak Bhattacharyya. 2023.
\newblock \href {https://doi.org/10.1145/3583780.3615265} {Multi-step prompting
  for few-shot emotion-grounded conversations}.
\newblock In \emph{Proceedings of the 32nd ACM International Conference on
  Information and Knowledge Management}, CIKM '23, page 3886–3891, New York,
  NY, USA. Association for Computing Machinery.

\bibitem[{Gao et~al.(2024)Gao, Zhang, Ding, and Zhao}]{10547289}
Yinfeng Gao, Qichao Zhang, Da-Wei Ding, and Dongbin Zhao. 2024.
\newblock \href {https://doi.org/10.1109/TIV.2024.3408830} {Dream to drive with
  predictive individual world model}.
\newblock \emph{IEEE Transactions on Intelligent Vehicles}, pages 1--16.

\bibitem[{Ha and Schmidhuber(2018)}]{haWorldModels2018}
David Ha and J{\"u}rgen Schmidhuber. 2018.
\newblock \href {https://doi.org/10.5281/zenodo.1207631} {World {{Models}}}.

\bibitem[{Hafner et~al.(2020)Hafner, Lillicrap, Ba, and
  Norouzi}]{Hafner2020Dream}
Danijar Hafner, Timothy Lillicrap, Jimmy Ba, and Mohammad Norouzi. 2020.
\newblock \href {https://openreview.net/forum?id=S1lOTC4tDS} {Dream to control:
  Learning behaviors by latent imagination}.
\newblock In \emph{International Conference on Learning Representations}.

\bibitem[{Hafner et~al.(2019)Hafner, Lillicrap, Fischer, Villegas, Ha, Lee, and
  Davidson}]{pmlr-v97-hafner19a}
Danijar Hafner, Timothy Lillicrap, Ian Fischer, Ruben Villegas, David Ha,
  Honglak Lee, and James Davidson. 2019.
\newblock \href {https://proceedings.mlr.press/v97/hafner19a.html} {Learning
  latent dynamics for planning from pixels}.
\newblock In \emph{Proceedings of the 36th International Conference on Machine
  Learning}, volume~97 of \emph{Proceedings of Machine Learning Research},
  pages 2555--2565. PMLR.

\bibitem[{Hu et~al.(2024)Hu, Wu, Zhu, Xianyu, Wang, Zhang, and
  Cao}]{hu2024openrlhf}
Jian Hu, Xibin Wu, Zilin Zhu, Xianyu, Weixun Wang, Dehao Zhang, and Yu~Cao.
  2024.
\newblock Openrlhf: An easy-to-use, scalable and high-performance rlhf
  framework.
\newblock \emph{arXiv preprint arXiv:2405.11143}.

\bibitem[{Jordan et~al.(1999)Jordan, Ghahramani, Jaakkola, and
  Saul}]{jordan1999viintro}
Michael~I Jordan, Zoubin Ghahramani, Tommi~S Jaakkola, and Lawrence~K Saul.
  1999.
\newblock An introduction to variational methods for graphical models.
\newblock \emph{Machine learning}, 37(2):183--233.

\bibitem[{Kang et~al.(2024)Kang, Kim, Kwon, Moon, Cho, Yu, Lee, and
  Yeo}]{kang-etal-2024-large}
Dongjin Kang, Sunghwan Kim, Taeyoon Kwon, Seungjun Moon, Hyunsouk Cho, Youngjae
  Yu, Dongha Lee, and Jinyoung Yeo. 2024.
\newblock \href {https://doi.org/10.18653/v1/2024.acl-long.813} {Can large
  language models be good emotional supporter? mitigating preference bias on
  emotional support conversation}.
\newblock In \emph{Proceedings of the 62nd Annual Meeting of the Association
  for Computational Linguistics (Volume 1: Long Papers)}, pages 15232--15261,
  Bangkok, Thailand. Association for Computational Linguistics.

\bibitem[{Li et~al.(2015)Li, Galley, Brockett, Gao, and
  Dolan}]{li2015diversity}
Jiwei Li, Michel Galley, Chris Brockett, Jianfeng Gao, and Bill Dolan. 2015.
\newblock A diversity-promoting objective function for neural conversation
  models.
\newblock \emph{arXiv preprint arXiv:1510.03055}.

\bibitem[{Li et~al.(2017)Li, Su, Shen, Li, Cao, and
  Niu}]{li-etal-2017-dailydialog}
Yanran Li, Hui Su, Xiaoyu Shen, Wenjie Li, Ziqiang Cao, and Shuzi Niu. 2017.
\newblock \href {https://aclanthology.org/I17-1099/} {{D}aily{D}ialog: A
  manually labelled multi-turn dialogue dataset}.
\newblock In \emph{Proceedings of the Eighth International Joint Conference on
  Natural Language Processing (Volume 1: Long Papers)}, pages 986--995, Taipei,
  Taiwan. Asian Federation of Natural Language Processing.

\bibitem[{Lin(2004)}]{lin2004rouge}
Chin-Yew Lin. 2004.
\newblock Rouge: A package for automatic evaluation of summaries.
\newblock In \emph{Text summarization branches out}, pages 74--81.

\bibitem[{Lin et~al.(2019)Lin, Madotto, Shin, Xu, and
  Fung}]{lin-etal-2019-moel}
Zhaojiang Lin, Andrea Madotto, Jamin Shin, Peng Xu, and Pascale Fung. 2019.
\newblock \href {https://doi.org/10.18653/v1/D19-1012} {{M}o{EL}: Mixture of
  empathetic listeners}.
\newblock In \emph{Proceedings of the 2019 Conference on Empirical Methods in
  Natural Language Processing and the 9th International Joint Conference on
  Natural Language Processing (EMNLP-IJCNLP)}, pages 121--132, Hong Kong,
  China. Association for Computational Linguistics.

\bibitem[{Liu et~al.(2021)Liu, Zheng, Demasi, Sabour, Li, Yu, Jiang, and
  Huang}]{liu2021ESconv}
Siyang Liu, Chujie Zheng, Orianna Demasi, Sahand Sabour, Yu~Li, Zhou Yu, Yong
  Jiang, and Minlie Huang. 2021.
\newblock \href {https://doi.org/10.18653/v1/2021.acl-long.269} {Towards
  emotional support dialog systems}.
\newblock In \emph{Proceedings of the 59th Annual Meeting of the Association
  for Computational Linguistics and the 11th International Joint Conference on
  Natural Language Processing (Volume 1: Long Papers)}, pages 3469--3483,
  Online. Association for Computational Linguistics.

\bibitem[{Maas et~al.(2011)Maas, Daly, Pham, Huang, Ng, and
  Potts}]{maas-EtAl:2011:ACL-HLT2011}
Andrew~L. Maas, Raymond~E. Daly, Peter~T. Pham, Dan Huang, Andrew~Y. Ng, and
  Christopher Potts. 2011.
\newblock \href {http://www.aclweb.org/anthology/P11-1015} {Learning word
  vectors for sentiment analysis}.
\newblock In \emph{Proceedings of the 49th Annual Meeting of the Association
  for Computational Linguistics: Human Language Technologies}, pages 142--150,
  Portland, Oregon, USA. Association for Computational Linguistics.

\bibitem[{Madaan et~al.(2023)Madaan, Tandon, Gupta, Hallinan, Gao, Wiegreffe,
  Alon, Dziri, Prabhumoye, Yang, Welleck, Majumder, Gupta, Yazdanbakhsh, and
  Clark}]{Madaan2023SelfRefine}
Aman Madaan, Niket Tandon, Prakhar Gupta, Skyler Hallinan, Luyu Gao, Sarah
  Wiegreffe, Uri Alon, Nouha Dziri, Shrimai Prabhumoye, Yiming Yang, Sean
  Welleck, Bodhisattwa~Prasad Majumder, Shashank Gupta, Amir Yazdanbakhsh, and
  Peter Clark. 2023.
\newblock \href {https://api.semanticscholar.org/CorpusID:257900871}
  {Self-refine: Iterative refinement with self-feedback}.
\newblock \emph{ArXiv}, abs/2303.17651.

\bibitem[{Mohammad and
  Kiritchenko(2018)}]{mohammad-kiritchenko-2018-understanding}
Saif Mohammad and Svetlana Kiritchenko. 2018.
\newblock \href {https://aclanthology.org/L18-1030/} {Understanding emotions: A
  dataset of tweets to study interactions between affect categories}.
\newblock In \emph{Proceedings of the Eleventh International Conference on
  Language Resources and Evaluation ({LREC} 2018)}, Miyazaki, Japan. European
  Language Resources Association (ELRA).

\bibitem[{Newman(2023)}]{Newman2023Efficient_BT}
M.~E.~J. Newman. 2023.
\newblock \href {http://jmlr.org/papers/v24/22-1086.html} {Efficient
  computation of rankings from pairwise comparisons}.
\newblock \emph{Journal of Machine Learning Research}, 24(238):1--25.

\bibitem[{Ouyang et~al.(2022)Ouyang, Wu, Jiang, Almeida, Wainwright, Mishkin,
  Zhang, Agarwal, Slama, Ray, Schulman, Hilton, Kelton, Miller, Simens, Askell,
  Welinder, Christiano, Leike, and Lowe}]{ouyangTrainingLanguageModels2022a}
Long Ouyang, Jeff Wu, Xu~Jiang, Diogo Almeida, Carroll~L. Wainwright, Pamela
  Mishkin, Chong Zhang, Sandhini Agarwal, Katarina Slama, Alex Ray, John
  Schulman, Jacob Hilton, Fraser Kelton, Luke Miller, Maddie Simens, Amanda
  Askell, Peter Welinder, Paul Christiano, Jan Leike, and Ryan Lowe. 2022.
\newblock \href {https://doi.org/10.48550/arXiv.2203.02155} {Training language
  models to follow instructions with human feedback}.
\newblock \emph{Preprint}, arXiv:2203.02155.

\bibitem[{Papineni et~al.(2002)Papineni, Roukos, Ward, and
  Zhu}]{papineni2002bleu}
Kishore Papineni, Salim Roukos, Todd Ward, and Wei-Jing Zhu. 2002.
\newblock Bleu: a method for automatic evaluation of machine translation.
\newblock In \emph{Proceedings of the 40th annual meeting of the Association
  for Computational Linguistics}, pages 311--318.

\bibitem[{Peng et~al.(2018)Peng, Li, Gao, Liu, and Wong}]{peng-etal-2018-deep}
Baolin Peng, Xiujun Li, Jianfeng Gao, Jingjing Liu, and Kam-Fai Wong. 2018.
\newblock \href {https://doi.org/10.18653/v1/P18-1203} {{D}eep {D}yna-{Q}:
  Integrating planning for task-completion dialogue policy learning}.
\newblock In \emph{Proceedings of the 56th Annual Meeting of the Association
  for Computational Linguistics (Volume 1: Long Papers)}, pages 2182--2192,
  Melbourne, Australia. Association for Computational Linguistics.

\bibitem[{Qian et~al.(2023)Qian, Wang, Ma, Bin, Zhang, Zhao, Huang, and
  Hou}]{10.5555/3545946.3598705}
Yushan Qian, Bo~Wang, Shangzhao Ma, Wu~Bin, Shuo Zhang, Dongming Zhao, Kun
  Huang, and Yuexian Hou. 2023.
\newblock Think twice: A human-like two-stage conversational agent for
  emotional response generation.
\newblock In \emph{Proceedings of the 2023 International Conference on
  Autonomous Agents and Multiagent Systems}, AAMAS '23, page 727–736,
  Richland, SC. International Foundation for Autonomous Agents and Multiagent
  Systems.

\bibitem[{Rafailov et~al.(2023)Rafailov, Sharma, Mitchell, Manning, Ermon, and
  Finn}]{Rafailov2023DPOLMisReward}
Rafael Rafailov, Archit Sharma, Eric Mitchell, Christopher~D Manning, Stefano
  Ermon, and Chelsea Finn. 2023.
\newblock Direct preference optimization: Your language model is secretly a
  reward model.
\newblock In \emph{Advances in Neural Information Processing Systems},
  volume~36, pages 53728--53741. Curran Associates, Inc.

\bibitem[{Rashkin et~al.(2019)Rashkin, Smith, Li, and
  Boureau}]{rashkin-etal-2019-towards}
Hannah Rashkin, Eric~Michael Smith, Margaret Li, and Y-Lan Boureau. 2019.
\newblock \href {https://doi.org/10.18653/v1/P19-1534} {Towards empathetic
  open-domain conversation models: A new benchmark and dataset}.
\newblock In \emph{Proceedings of the 57th Annual Meeting of the Association
  for Computational Linguistics}, pages 5370--5381, Florence, Italy.
  Association for Computational Linguistics.

\bibitem[{Schulman et~al.(2017)Schulman, Wolski, Dhariwal, Radford, and
  Klimov}]{schulman2017proximal}
John Schulman, Filip Wolski, Prafulla Dhariwal, Alec Radford, and Oleg Klimov.
  2017.
\newblock Proximal policy optimization algorithms.
\newblock \emph{arXiv preprint arXiv:1707.06347}.

\bibitem[{Socher et~al.(2013)Socher, Perelygin, Wu, Chuang, Manning, Ng, and
  Potts}]{socher-etal-2013-recursive}
Richard Socher, Alex Perelygin, Jean Wu, Jason Chuang, Christopher~D. Manning,
  Andrew Ng, and Christopher Potts. 2013.
\newblock \href {https://aclanthology.org/D13-1170/} {Recursive deep models for
  semantic compositionality over a sentiment treebank}.
\newblock In \emph{Proceedings of the 2013 Conference on Empirical Methods in
  Natural Language Processing}, pages 1631--1642, Seattle, Washington, USA.
  Association for Computational Linguistics.

\bibitem[{Sutton(1991)}]{suttonDynaIntegratedArchitecture1991}
Richard~S. Sutton. 1991.
\newblock \href {https://doi.org/10.1145/122344.122377} {Dyna, an integrated
  architecture for learning, planning, and reacting}.
\newblock \emph{SIGART Bull.}, 2(4):160--163.

\bibitem[{Tishby et~al.(2000)Tishby, Pereira, and Bialek}]{tishby2000ib}
Naftali Tishby, Fernando~C Pereira, and William Bialek. 2000.
\newblock The information bottleneck method.
\newblock \emph{arXiv preprint physics/0004057}.

\bibitem[{Wang et~al.(2024{\natexlab{a}})Wang, Deng, Lyu, Zeng, He, Yan, and
  An}]{wangImprovingMultistepReasoning2024b}
Chaojie Wang, Yanchen Deng, Zhiyi Lyu, Liang Zeng, Jujie He, Shuicheng Yan, and
  Bo~An. 2024{\natexlab{a}}.
\newblock \href {https://doi.org/10.48550/arXiv.2406.14283} {Q*: {{Improving
  Multi-step Reasoning}} for {{LLMs}} with {{Deliberative Planning}}}.
\newblock \emph{Preprint}, arXiv:2406.14283.

\bibitem[{Wang et~al.(2024{\natexlab{b}})Wang, Yang, Huang, Yang, Majumder, and
  Wei}]{wang2024E5}
Liang Wang, Nan Yang, Xiaolong Huang, Linjun Yang, Rangan Majumder, and Furu
  Wei. 2024{\natexlab{b}}.
\newblock \href {https://doi.org/10.18653/v1/2024.acl-long.642} {Improving text
  embeddings with large language models}.
\newblock In \emph{Proceedings of the 62nd Annual Meeting of the Association
  for Computational Linguistics (Volume 1: Long Papers)}, pages 11897--11916,
  Bangkok, Thailand. Association for Computational Linguistics.

\bibitem[{Wang et~al.(2024{\natexlab{c}})Wang, He, {yang}, Wang, Meng, Pan, and
  Sui}]{wangFSMFiniteState2024}
Xiaochen Wang, Junqing He, Zhe {yang}, Yiru Wang, Xiangdi Meng, Kunhao Pan, and
  Zhifang Sui. 2024{\natexlab{c}}.
\newblock \href {https://doi.org/10.48550/arXiv.2407.02964} {{{FSM}}: {{A
  Finite State Machine Based Zero-Shot Prompting Paradigm}} for {{Multi-Hop
  Question Answering}}}.
\newblock \emph{Preprint}, arXiv:2407.02964.

\bibitem[{Wang and Zhao(2024)}]{wang-zhao-2024-metacognitive}
Yuqing Wang and Yun Zhao. 2024.
\newblock \href {https://doi.org/10.18653/v1/2024.naacl-long.106}
  {Metacognitive prompting improves understanding in large language models}.
\newblock In \emph{Proceedings of the 2024 Conference of the North American
  Chapter of the Association for Computational Linguistics: Human Language
  Technologies (Volume 1: Long Papers)}, pages 1914--1926, Mexico City, Mexico.
  Association for Computational Linguistics.

\bibitem[{Wei et~al.(2022)Wei, Wang, Schuurmans, Bosma, Xia, Chi, Le, Zhou
  et~al.}]{wei2022chain}
Jason Wei, Xuezhi Wang, Dale Schuurmans, Maarten Bosma, Fei Xia, Ed~Chi, Quoc~V
  Le, Denny Zhou, et~al. 2022.
\newblock Chain-of-thought prompting elicits reasoning in large language
  models.
\newblock \emph{Advances in Neural Information Processing Systems},
  35:24824--24837.

\bibitem[{Wu et~al.(2021)Wu, Fang, Wang, Cao, Bao, Ping, Zhu, and
  Wang}]{wu-etal-2021-gaussian}
Guanlin Wu, Wenqi Fang, Ji~Wang, Jiang Cao, Weidong Bao, Yang Ping, Xiaomin
  Zhu, and Zheng Wang. 2021.
\newblock \href {https://doi.org/10.18653/v1/2021.findings-acl.156} {{G}aussian
  process based deep {D}yna-{Q} approach for dialogue policy learning}.
\newblock In \emph{Findings of the Association for Computational Linguistics:
  ACL-IJCNLP 2021}, pages 1786--1795, Online. Association for Computational
  Linguistics.

\bibitem[{Xu et~al.(2025)Xu, Wang, Zhao, and Fang}]{xu-etal-2025-efficient}
Kai Xu, Zhenyu Wang, Yangyang Zhao, and Bopeng Fang. 2025.
\newblock \href {https://aclanthology.org/2025.coling-main.490/} {An efficient
  dialogue policy agent with model-based causal reinforcement learning}.
\newblock In \emph{Proceedings of the 31st International Conference on
  Computational Linguistics}, pages 7331--7343, Abu Dhabi, UAE. Association for
  Computational Linguistics.

\bibitem[{Young et~al.(2013)Young, Gašić, Thomson, and Williams}]{6407655}
Steve Young, Milica Gašić, Blaise Thomson, and Jason~D. Williams. 2013.
\newblock \href {https://doi.org/10.1109/JPROC.2012.2225812} {Pomdp-based
  statistical spoken dialog systems: A review}.
\newblock \emph{Proceedings of the IEEE}, 101(5):1160--1179.

\bibitem[{Zhou et~al.(2024{\natexlab{a}})Zhou, Pang, Shen, and
  Cheng}]{zhou-etal-2024-think}
Junkai Zhou, Liang Pang, Huawei Shen, and Xueqi Cheng. 2024{\natexlab{a}}.
\newblock \href {https://doi.org/10.18653/v1/2024.findings-naacl.248} {Think
  before you speak: Cultivating communication skills of large language models
  via inner monologue}.
\newblock In \emph{Findings of the Association for Computational Linguistics:
  NAACL 2024}, pages 3925--3951, Mexico City, Mexico. Association for
  Computational Linguistics.

\bibitem[{Zhou et~al.(2024{\natexlab{b}})Zhou, Zanette, Pan, Levine, and
  Kumar}]{zhouArCHerTrainingLanguage2024b}
Yifei Zhou, Andrea Zanette, Jiayi Pan, Sergey Levine, and Aviral Kumar.
  2024{\natexlab{b}}.
\newblock \href {https://doi.org/10.48550/arXiv.2402.19446} {{{ArCHer}}:
  {{Training Language Model Agents}} via {{Hierarchical Multi-Turn RL}}}.
\newblock \emph{Preprint}, arXiv:2402.19446.

\end{thebibliography}
